\documentclass{article}
\usepackage[preprint]{colm2025_conference}

\usepackage{microtype}
\usepackage[utf8]{inputenc} 
\usepackage[T1]{fontenc}    
\definecolor{green}{RGB}{0,150,10}
\definecolor{blue}{RGB}{0,148,181}
\definecolor{orange}{RGB}{194,153,107}
\usepackage[colorlinks,linkcolor=red,anchorcolor=green,citecolor=blue]{hyperref}
\usepackage{url}            

\usepackage{lineno}
\usepackage{adjustbox}
\usepackage{booktabs}       
\usepackage{amsfonts}       
\usepackage{nicefrac}       
\usepackage{xcolor}         
\usepackage{amsmath}
\usepackage{marvosym}
\usepackage{graphicx}
\usepackage{colortbl}
\usepackage{multirow}
\usepackage{subcaption}
\usepackage{changes}
\usepackage{listings}
\usepackage{mdframed}
\usepackage{caption}
\usepackage[most]{tcolorbox}
\usepackage{enumitem}
\usepackage{tcolorbox}
\usepackage{geometry}

\newcommand{\ieno}{\textit{i.e.}}
\newcommand{\egno}{\textit{e.g.}}
\newcommand{\tct}{\textcolor{black}}

\title{\textsc{OmniCaptioner}: One Captioner to Rule Them All}

\newcommand{\homepage}{\raisebox{-1.5pt}{\includegraphics[height=1em]{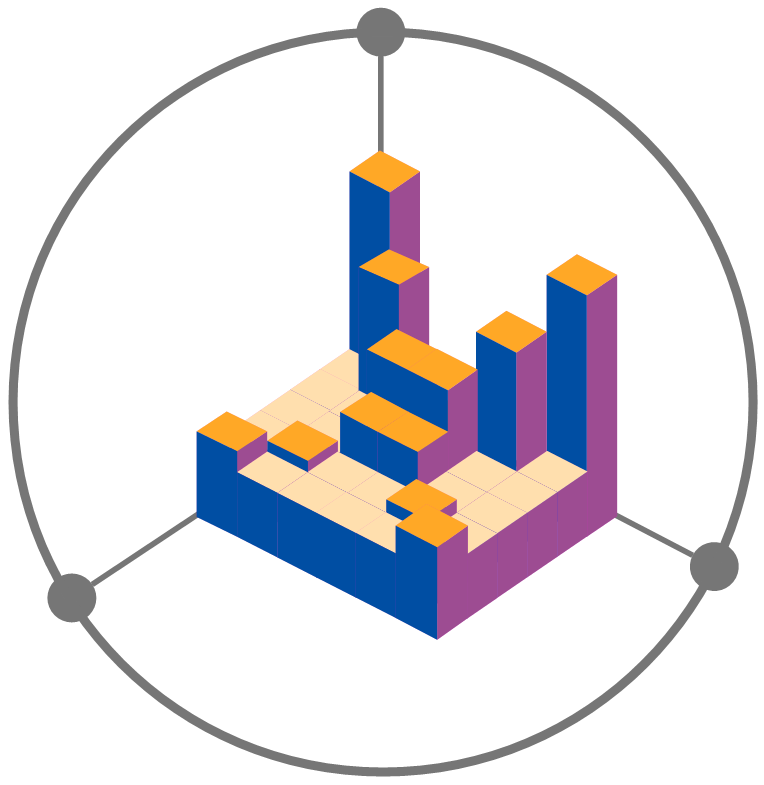}}}
\newcommand{\github}{\raisebox{-1.5pt}{\includegraphics[height=1em]{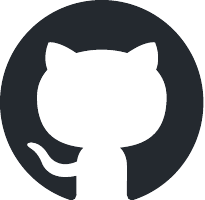}}}
\newcommand{\huggingface}{\raisebox{-1.5pt}{\includegraphics[height=1em]{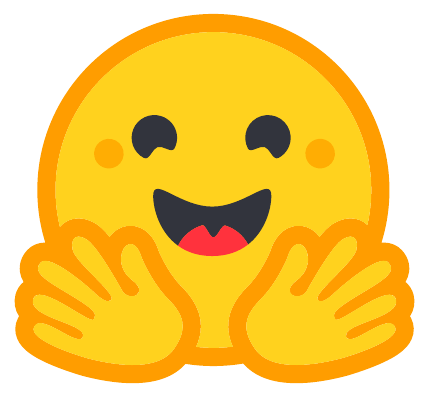}}}
\renewcommand{\thefootnote}{\fnsymbol{footnote}}
\author{%
  Yiting Lu$^{2,\dag}$, Jiakang Yuan$^{1,3,\dag}$, Zhen Li$^{4}$, Shitian Zhao$^{1}$, Qi Qin$^{1}$, Xinyue Li$^{1}$, Le Zhuo$^{1}$ \\ \bf Licheng Wen$^{1}$, Dongyang Liu$^{1}$, Yuewen Cao$^{1}$, Xiangchao Yan$^{1}$, Xin Li$^{2}$, Tianshuo Peng$^{1,4}$ \\ \bf  Shufei Zhang$^{1}$, Botian Shi$^{1}$, Tao Chen$^{3}$, Zhibo Chen$^{2,\text{\Letter}}$, Lei Bai$^{1}$, Peng Gao$^{1}$, Bo Zhang$^{1,\ddagger,\text{\Letter}}$ \\ [1mm]
\textsuperscript{\rm 1} Shanghai Artificial Intelligence Laboratory,~~
\textsuperscript{\rm 2} University of Science and Technology of China, \\
\textsuperscript{\rm 3} Fudan University, \textsuperscript{\rm 4} The Chinese University of Hong Kong\\ [1.5mm]
{\homepage\ \texttt{\url{https://alpha-innovator.github.io/OmniCaptioner-project-page}}} \\
{\github\ \texttt{\url{https://github.com/Alpha-Innovator/OmniCaptioner}}} \\
  {\huggingface\ \texttt{\url{https://huggingface.co/U4R/OmniCaptioner}}}
}

\begin{document}

\maketitle

\renewcommand{\thefootnote}{}
\footnotetext{$^{\dag}$ Equal contribution, $^{\ddagger}$~Project Lead, $^{\textrm{\Letter}}$ Corresponding authors.}

\begin{figure}[h]
\vspace{-10pt}
    \centering
    \includegraphics[width=\linewidth]{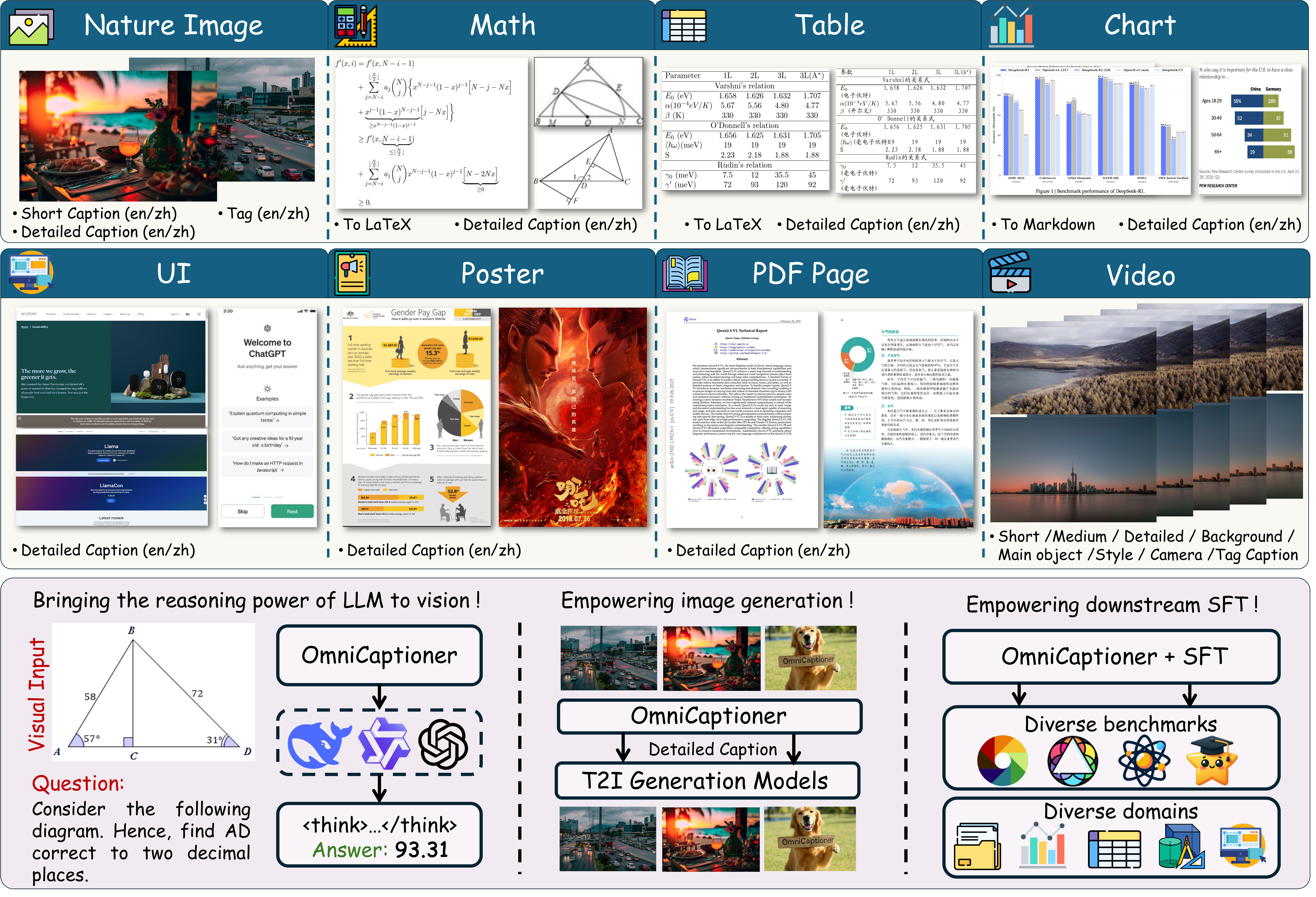}
    \vspace{-18pt}\caption{\textsc{\textbf{OmniCaptioner}}: the top section demonstrates its capability to process diverse visual domains. The bottom section highlights its applications in visual reasoning (associated with reasoning LLM), image generation (integrated with T2I generation models), and efficient downstream SFT tasks adaptation.}
    \label{fig:figure1}
\end{figure}

\begin{abstract}
We propose \textsc{OmniCaptioner}, a versatile visual captioning framework for generating fine-grained textual descriptions across a wide variety of visual domains. Unlike prior methods limited to specific image types (\egno, natural images or geometric visuals), our framework provides a unified solution for captioning natural images, visual text (\egno, posters, UIs, textbooks), and structured visuals (\egno, documents, tables, charts). By converting low-level pixel information into semantically rich textual representations, our framework bridges the gap between visual and textual modalities. Our results highlight three key advantages: (i) Enhanced Visual Reasoning with LLMs, where long-context captions of visual modalities empower LLMs, particularly the DeepSeek-R1 series, to reason effectively in multimodal scenarios; (ii) Improved Image Generation, where detailed captions improve tasks like text-to-image generation and image transformation; and (iii) Efficient Supervised Fine-Tuning (SFT), which enables faster convergence with less data. We believe the versatility and adaptability of \textsc{OmniCaptioner} can offer a new perspective for bridging the gap between language and visual modalities.

\end{abstract}

\section{Introduction}
\label{sec:intro}

Pretraining of Multimodal Large Language Models (MLLMs)~\citep{llava,Llava-onevision,internvl2,qwen2vl,bai2025qwen2.5-vl}, particularly in bridging the gap between visual and textual domains, has gained significant attention in recent years. Substantial progress has been achieved in image captioning and visual question answering, enabling models to serve as universal visual assistants through large-scale Supervised Fine-Tuning (SFT). However, MLLMs still face limitations in perceptual accuracy in the visual-text and structured image domains, particularly when handling synthesized images that exhibit a substantial domain gap from natural images, as illustrated in Fig.~\ref{fig:compare_caption} (c).

\begin{figure}[tbp]
  \centering
   \includegraphics[width=1.0\linewidth]{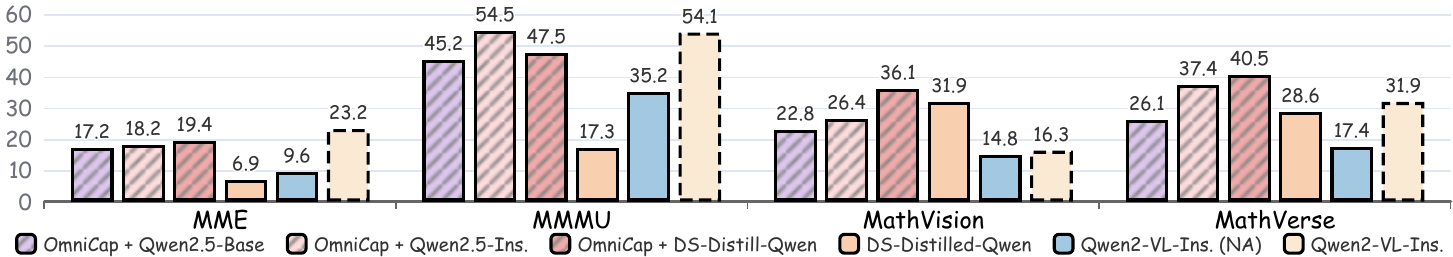}
   \caption{Performance comparison across different visual benchmarks for different LLMs/MLLMs (7B) with or without visual input. The bar with dashed borders denotes Qwen2-VL-Instruct, indicating it has pixel-level visual input, while others do not. Qwen2-VL-Ins.(NA) refers to a setting where only the question is provided as input. We divide the MME score by 100 to have the same scale as other benchmarks.}
   \label{fig:wovisualinput}
\end{figure}

\begin{figure}[tbp]
    \centering
    \includegraphics[width=0.95\textwidth]{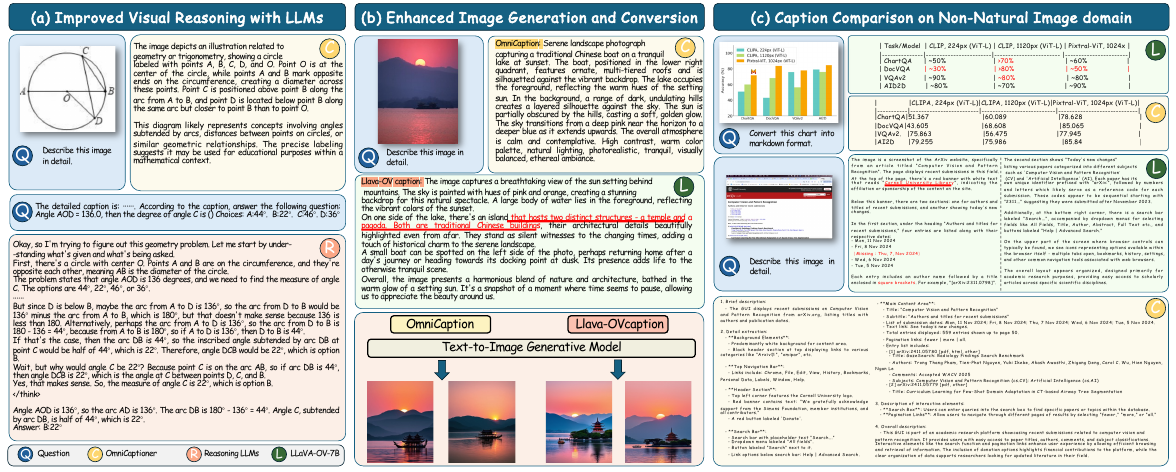}
    \caption{Illustration of \textsc{OmniCaptioner}’s plug-and-play applications (Sub-figure a, b) and comparison between \textsc{OmniCaptioner} and LLava-OneVision-7B on non-natural image captioning (Sub-figure c). Sub-figure (a) shows that \textsc{OmniCaptioner} leverages LLMs' strong reasoning abilities to perform multimodal reasoning tasks. Sub-figure (b) highlights how hallucinated or inaccurate captions—like those from LLava-OneVision-7B can lead to inconsistent image conversion, revealing weakened alignment capabilities in text-to-image models when captions don't faithfully represent the original content. Sub-figure (c) highlights that LLaVA-OneVision-7B, due to limited exposure to non-natural images during pretraining, struggles with perception in such domains, often leading to hallucinations, whereas \textsc{OmniCaptioner} provides more accurate descriptions.}

    \label{fig:compare_caption}
\end{figure}

Recent research has increasingly emphasized the role of image captioning in aligning modalities during multimodal pretraining, aiming to enhance both perception and reasoning across diverse domains through the SFT process. Meanwhile, domain-specific studies, such as those focusing on document understanding MLLMs~\citep{layoutllm, mplug-docowl-1.5} and mathematical MLLMs~\citep{peng2024chimera, zhang2025open, xia2024geox}, have leveraged domain-specific caption data to further improve modality alignment and advance multimodal pretraining.
\textit{These advancements highlight the need for a unified framework for multimodal pretraining centered on image captioning.} Also, despite progress in MLLMs, their multimodal reasoning capabilities still fall short of the textual reasoning abilities of LLMs. As shown in Fig.~\ref{fig:wovisualinput}, when provided only with a question and no visual input on the MathVision and MathVerse benchmarks, DeepSeek-Distill-Qwen-7B (orange) significantly outperforms Qwen2-VL-Instruct (blue), demonstrating the strength of LLM-driven reasoning in multimodal tasks.

In this work, we bridge this gap by introducing the first \textsc{OmniCaptioner} framework, designed to generate fine-grained textual descriptions across diverse visual domains as shown in Fig.~\ref{fig:figure1}. Unlike prior approaches that focus on specific visual categories (\ieno, natural or geometry images), our approach enables a unified solution for diverse image types, paving the way for broader multimodal understanding. We focus on converting low-level pixel features into semantically rich textual representations, which preserve crucial visual details while bridging the modality gap between vision and language. \textsc{OmniCaptioner} has two characteristics:
i) \textbf{Diverse Visual Domain Coverage}: We present a unified framework that supports diverse visual content, including natural images, visual text images (\egno, poster, UI, textbook) and structured images (\egno, geometry, equation, tables, charts).
ii) \textbf{Pixel-to-Text Mapping}: By pairing these diverse image types with detailed captions, we convert low-level pixel information into semantically rich, fine-grained textual descriptions, enabling a deeper understanding of visual content, which effectively bridges the gap between visual and textual modalities.

To evaluate the effectiveness of \textsc{OmniCaptioner}, we conduct systematic assessments across both image understanding (\egno, visual reasoning) and image generation tasks (\egno, text-to-image generation). Our results reveal several key advantages:
i) \textbf{Improved Visual Reasoning with LLMs}: Our detailed, long-context captions can be directly incorporated into powerful LLMs to address challenging visual reasoning questions, particularly for models like the DeepSeek-R1~\citep{Deepseek-r1} series. This approach enables LLMs to perform visual reasoning tasks in a training-free manner, leveraging rich textual descriptions without requiring additional fine-tuning.
ii) \textbf{Enhanced Image Generation and Conversion}: The detailed captions produced by our framework significantly improve image generation tasks, such as image-to-text generation and image conversion, owing to their near-complete pixel-to-text mapping capability.
iii) \textbf{Efficient SFT Process}: Leveraging the knowledge from pretraining on \textsc{OmniCaptioner}, the SFT process becomes more efficient, requiring less training data and achieving faster convergence.

Furthermore, the contributions of this paper are summarized below:

\begin{itemize}
    \item \textbf{Unified Visual Captioning Framework}: 
    We present \textsc{OmniCaptioner}, a unified framework for generating captions across diverse domains. Our approach seamlessly integrates captioning capabilities for natural images, visual text images (\egno, posters, UI, and textbooks), and structured visual images (\egno, tables, charts, equations, and geometric diagrams).  \textsc{OmniCaptioner} sets a new standard for generalized visual captioning, enabling more effective and scalable vision-language understanding.
    \item \textbf{Comprehensive Pixel-to-Text Conversion}: Our framework leverages detailed captions to convert low-level pixel information into semantically rich, fine-grained textual descriptions, effectively bridging the gap between visual and textual modalities. Particularly, this enhances text-to-image generation by providing more precise and context-aware textual guidance, leading to improved visual fidelity and alignment with the intended semantics.
    \item \textbf{Improved Visual Reasoning with LLMs}: By incorporating detailed, long-context captions, our approach enables enhanced visual reasoning capabilities, especially when integrated into LLMs such as the DeepSeek-R1 series.  Leveraging the perceptual information provided by \textsc{OmniCaptioner}, LLMs can infer and reason within the textual space to effectively solve visual reasoning tasks.
\end{itemize}

\section{Related Works}
\noindent \textbf{Image Captioning.} Image captioning tasks can be broadly classified into two categories. The first approach focuses on generating high-quality captions for natural images. Notably, ShareGPT4V~\citep{chen2024sharegpt4v} improves vision-language alignment by collecting high-quality, attribute-specific captions through targeted prompts to GPT-4V for natural images, while models like Densefusion~\citep{li2024densefusion} leverage multiple expert models to synthesize captions for natural images. The second approach, exemplified by  CompCap~\citep{chen2024compcap}, tackles the challenge of domain diversity during pretraining by incorporating synthetic images to enhance performance on underrepresented domains. However, the first approaches are often constrained by its focus on specific domains, while the second faces challenges due to the relatively small quantity of synthetic images used during training.

\noindent \textbf{Multimodal Large Language Models. }
With the development of LLMs~\citep{yang2024qwen2-5,Deepseek-r1,touvron2023llama,yuan2025dolphin}, integrating visual perception capability into LLMs (\ieno, MLLMs) has received increasing attention. To address the gap between different modalities, most of works~\citep{qwen2vl,bai2025qwen2.5-vl,internvl2,xia2024chartx,llava,Llava-onevision,lin2023sphinx,liu2024sphinx-x} first pretrain on image captioning data to obtain a vision-language connector (\egno, MLP-based or cross-attention based) and followed by SFT. To better integrate information from multiple modalities, several works~\citep{lin2024moma,luo2024mono-internvl,diao2024unveiling,team2024chameleon} try to explore new architectures to process different modalities in a single Transformer model. In addition to model architecture, some works~\citep{wang2024enhancing} try to boost models' reasoning ability through post-training (\egno, reinforcement learning)~\citep{wang2024enhancing} or test-time scaling (\egno, monte-carlo tree search)~\citep{yao2024mulberry,luo2025ursa,dong2024progressive}. Furthermore, recent studies~\citep{zhang2024mm1,mckinzie2024mm1,chen2024compcap,deng2025coconut} have systematically investigated the influence of data quality on on both the pretraining and SFT phases of MLLMs. MM1~\citep{zhang2024mm1} reveals that model capabilities induced through pretraining with high-quality data are effectively preserved after SFT. Most existing open-source MLLMs~\citep{llava,Llava-onevision} primarily focus on pretraining with natural images, while domain-specific MLLMs (\egno, math, chart) are trained exclusively on domain-specific caption data. In contrast, we propose a more unified pretraining approach that integrates diverse domain knowledge during pretraining. In addition, current MLLMs generally exhibit inferior reasoning capabilities compared to text-only LLMs, \textsc{OmniCaptioner} can generate detailed, long-context captions of different domains and use LLMs to address challenging visual reasoning tasks.

\section{\textsc{OmniCaptioner}}

\begin{figure*}[h]
  \centering
   \includegraphics[width=1.0\linewidth]{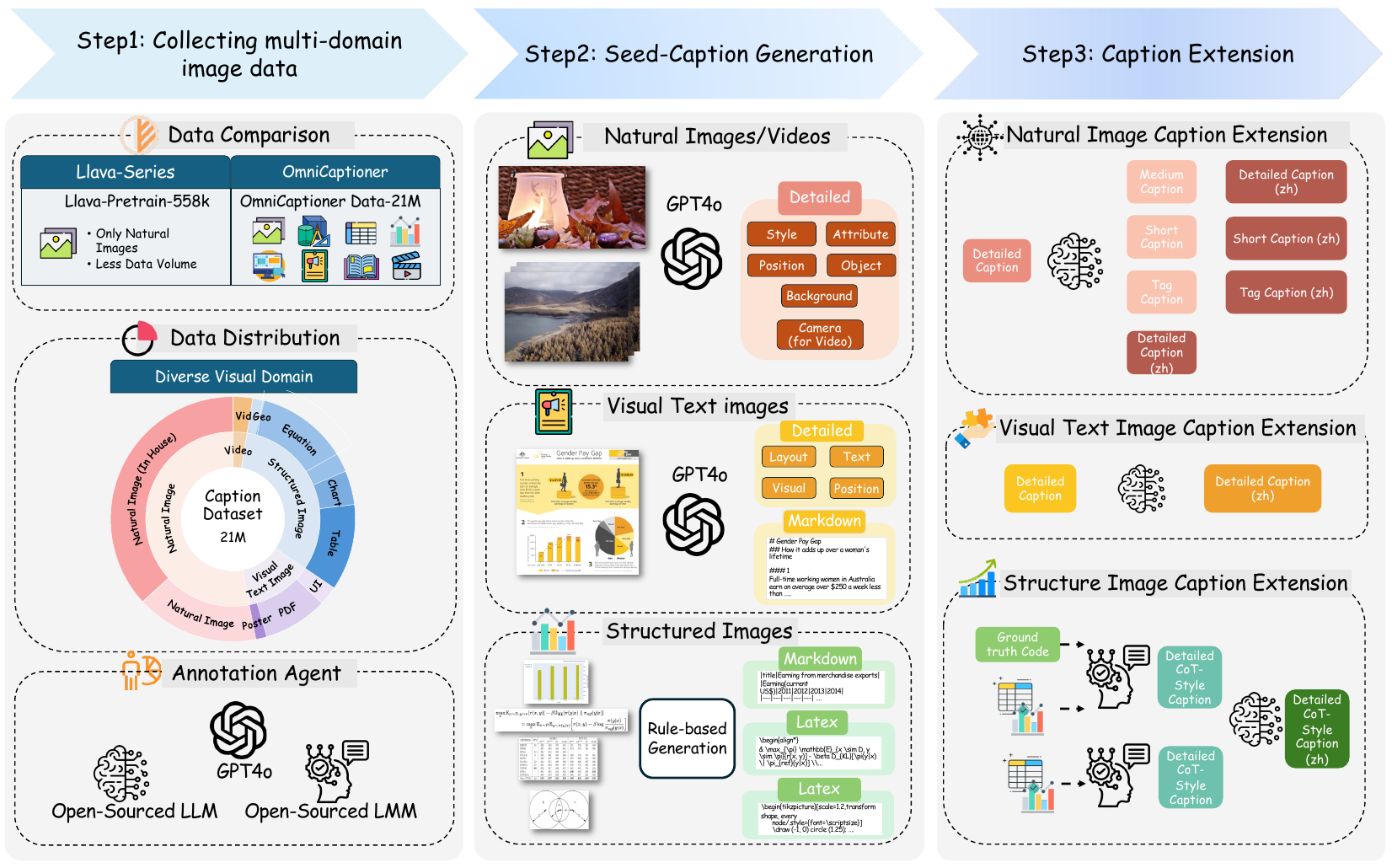}
   \caption{\textsc{OmniCaptioner}’s diverse visual captioning pipeline.
The pipeline consists of Seed-Caption Generation to ensure precise pixel-to-word mapping, and Caption Extension to enrich caption styles to support image generation and visual reasoning tasks. \textsc{OmniCaptioner} utilizes a \textbf{21M-caption dataset}, covering diverse domains beyond natural images, enabling more comprehensive captioning capabilities. For further details about dataset composition, please refer to Fig.~\ref{fig:data_src} in Appendix~\ref{sec:data_comp}.}
\label{fig:framework}
\end{figure*}

To achieve a unified multimodal pretraining paradigm and handle diverse visual domains, we first construct a diverse caption dataset as shown in Sec.~\ref{fig:framework}. We will provide the dataset description and then detail the dataset construction process in Sec.~\ref{sec:data_description} and Sec.~\ref{sec:data_construction}, respectively. And the pertaining process is described in  Sec.~\ref{sec:Pretraining}.

\subsection{Diverse Visual Caption Dataset}
\label{sec:data_description}

The diversity of our visual caption dataset is characterized by two dimensions: domain diversity (diverse data sources) and caption formula diversity. To achieve effective unified pretraining, the dataset needs to encompass a wider range of domains. For example, when acting as a documentation assistant, MLLMs need to comprehend tables and charts, while as a GUI agent, they are required to understand elements in web pages. As illustrated in the data distribution section of Fig.~\ref{fig:framework}, our caption dataset is composed of four major categories: natural images, structured images (including chart, table, and so on), visual text images (including UI images, posters, and so on), and video. This comprehensive data coverage enables our model to serve as a multi-domain assistant and further enhance the performance on downstream tasks. Furthermore, diverse types of captions may be necessary even for the same visual input. For instance, a chart image may require both structured tabular conversion and comprehensive analytical descriptions. To address this requirement, we define diverse caption formulas for each domain. This approach enables our model to generate diverse caption formats, including multilingual (Chinese and English) descriptions, varying granularity levels (from comprehensive to concise), and so on.

\subsection{Dataset Construction}
\label{sec:data_construction}
\vspace{-1mm}

To generate high-quality captions for images across diverse domains, we propose a two-step caption generation pipeline. The design of our pipeline takes into account the need for accurate visual descriptions, the flexibility to support different stylistic outputs, the ability to perform reasoning and logic extrapolation, as well as bilingual captioning.

\noindent \textbf{Seed Caption Generation.} In the first stage, we focus on seed caption generation. The goal is to produce an initial caption that is as accurate as possible, with a comprehensive textual description of all relevant visual elements present in the visual signal. 
This stage leverages carefully designed prompts to guide the powerful closed-source multimodal model GPT-4o to describe all possible visual elements in natural images and visual-text images, ensuring an accurate pixel-to-word mapping. For structured images generated via code, the description is generated as accurately as possible using predefined code rules.
The generated seed caption serves as a reliable foundation for further refinement in the subsequent stage.

\noindent\textbf{Caption Extension.}
The second stage, caption extension, is responsible for enhancing and diversifying the generated caption. Here, the focus shifts from purely accuracy to incorporating stylistic variation and domain-specific reasoning. The seed caption is extended by introducing bilingual outputs (Chinese and English), with variations ranging from detailed to medium-length, short, and tag-style captions. Additionally, we inject reasoning knowledge relevant to specific domains to enrich the semantic depth of the captions. This allows the captions to not only reflect the visual content but also accommodate nuanced understanding in different contexts. Specially, \textbf{for Natural Images}, we leverage the open-source LLM, Qwen2.5-32B, to adjust the caption length through different prompts, allowing captions to range from medium-length to short and tag-style. Additionally, these varied captions are translated into Chinese, facilitating the creation of bilingual prompts for image generation. The benefit of this approach is to enable more flexible and effective bilingual prompt extraction for image generation tasks.
\textbf{For Visual Text Images}, we use open source LLM Qwen2.5-32B to translate the detailed subtitles generated by GPT-4o into the corresponding Chinese versions to ensure cross-language consistency.
\textbf{For Structured Images}, which often relate to mathematical or document-based reasoning (\egno, Chain-of-Thought (CoT) analysis), we prioritize the accuracy of the seed caption. After confirming the seed caption’s accuracy, we input both the seed caption and the original image into the open-source multimodal model Qwen2-VL-76B for CoT-style caption generation. This approach allows us to condition the captioning process on both the seed caption’s code (\egno, Markdown, LaTeX) and the image content, reducing hallucinations and improving the reliability of the generated captions. \tct{Additionally, we collect structured images without seed captions and directly input them into the same multimodal model for CoT-style caption generation.}
By decoupling the caption generation process into these two stages, we ensure both high accuracy in representing visual content and flexibility in producing diverse, contextually appropriate captions. 

\subsection{Unified Pretraining Process}
\label{sec:Pretraining}
\vspace{-1mm}
To effectively handle the multi-domain nature of the \textsc{OmniCaptioner} dataset, which spans a broad range of image types and captioning tasks, we propose a practical approach utilizing distinct system prompts. These prompts help minimize task conflicts and improve task coordination during training. By customizing system prompts for specific image categories and using a fixed set of question templates for various captioning styles, we differentiate between tasks and data types in the pretraining process. This approach facilitates efficient multi-domain training, ensuring robust model performance across diverse tasks and domains. To address the challenge of handling images with large variations in resolution and arbitrary aspect ratios, we leverage the \tct{powerful visual understanding} capabilities of the Qwen2-VL-7B~\citep{qwen2vl} model. Given that the Qwen2-VL-Instruct model is inherently powerful in managing multi-domain image data, we initialize our model with the Qwen2-VL-Instruct weights. This initialization allows us to effectively fine-tune on the \textsc{OmniCaptioner} dataset, ensuring robust performance across a wide range of image resolutions and aspect ratios while benefiting from the model's ability to generalize across diverse domains.

\section{One Captioner to Rule Them All}

\textbf{Improved Visual Reasoning Tasks with LLMs.} \tct{Current MLLMs lag behind LLMs in reasoning capabilities. This discrepancy motivates us to investigate whether LLMs can directly perform visual reasoning without modality-alignment losses that may degrade their reasoning ability while still effectively handling diverse visual reasoning tasks.}
In this work, we integrate image captioning with large language models (LLMs) to enable seamless visual reasoning \tct{in textual space}. 
As illustrated in Fig.~\ref{fig:compare_caption} (a), firstly, our captioner converts input images (spanning natural images, charts, equations, and beyond) into linguistically dense descriptions that explicitly encode pixel-level structures (\egno, spatial layouts, symbolic operators, tabular hierarchies) into textual space. These captions, acting as lossless semantic proxies, are then directly processed by powerful LLMs (\egno, DeepSeek-R1~\citep{Deepseek-r1}, Qwen2.5 series~\citep{yang2024qwen2-5}) to perform task-agnostic visual reasoning, including geometric problem-solving and spatial analysis.

Just shown in Fig.~\ref{fig:compare_caption}, \textsc{OmniCaptioner} can transform geometric images into detailed and precise visual descriptions. \textsc{OmniCaptioner} accurately describes geometric images, such as a circle with a diameter and circumferential angles, detailing spatial relationships among points. This enables LLMs to perform logical inferences, like calculating angles, without direct pixel-level perception.

There are three key advantages to this approach: 
i) Decoupled Perception and Reasoning – By separating perception (handled by MLLMs) from reasoning (handled by LLMs), our method avoids conflicts between the two capabilities, leading to more effective and accurate visual reasoning.
ii) Elimination of Modality-Alignment Training – Instead of requiring complex modality-alignment losses, our approach translates visual inputs into linguistic representations, allowing LLMs to process them naturally. This removes the need for additional multimodal training while preserving the reasoning strengths of LLMs.
iii) Flexibility and Generalization – The plug-and-play design enables seamless integration of LLMs into diverse visual reasoning tasks without domain-specific tuning. This ensures broad applicability across different types of visual inputs, from geometric diagrams to complex tabular structures.


\noindent \textbf{Enhanced Image Generation and Conversion.} Detailed and accurate image captions play a pivotal role in both the training and inference stages of Text-to-Image (T2I) tasks. During training, such captions offer fine-grained supervision by explicitly aligning low-level/high-level visual patterns (\egno, textures, spatial arrangements, object attributes) with precise linguistic semantics. 
At inference time, as shown in Fig.~\ref{fig:compare_caption}(b), detailed and precise captions substantially enhance image generation quality by guiding the model to follow instructions more faithfully—capturing spatial relationships, object interactions, and semantic details with higher fidelity. These benefits highlight the critical role of captions as a dense supervisory signal, enabling more precise instruction-following in T2I generation.

\begin{table}[t]
\caption{Performance comparison on various visual benchmarks between our \textsc{OmniCaptioner}-inserted LLMs and previous SOTA MLLMs. \textbf{We would like to emphasize} that by utilizing \textsc{OmniCaptioner}, LLMs can function as MLLMs without requiring additional training. Moreover, we have observed that, particularly in mathematical scenarios, caption-integrated LLMs surpasses MLLMs with comparable parameter sizes, where MLLMs have undergone rigorous data preparation and GPU-intensive training.}
\label{caption_LLM}
\centering
\small
\resizebox{1.0\linewidth}{!}{
\begin{tabular}{l c c c c c}
\toprule
\textbf{Model}  & \textbf{MME} & \textbf{MMMU} & \textbf{MathVision} & \textbf{MathVerse} & \textbf{Olympiad} \\
\cmidrule{1-6}
\multicolumn{6}{c}{\textbf{\textit{Frontier Models}}} \\
GPT-4V  & - & 63.1 & 24.0 & 32.8&18.0 \\
GPT-4o (2024-05)  & - & 69.1 & 30.4 & 50.2 & 25.9 \\
Claude3.5-Sonnet & - & 68.3 & - & - & - \\

\cmidrule{1-6}
\multicolumn{6}{c}{\textit{\textbf{3B-Level Models}}} \\
Qwen2-VL-2B~\citep{qwen2vl} & 1872 & 41.1  & 12.4 & 21.0  & -  \\
InternVL2-2B~\citep{internvl2} &\textbf{1876}  &36.3  & 12.1 & \textbf{25.3}  & 0.4  \\
MinniCPM-V2.0~\citep{yao2024minicpm} & 1808  & 38.2 &-  & - & - \\
\rowcolor{gray!30}\textsc{OmniCaptioner} + Qwen2.5-3B-Instruct &1599 & \textbf{43.0} & \textbf{16.0} & 22.2 & \textbf{7.24} \\
\cmidrule{1-6}
\multicolumn{6}{c}{\textit{\textbf{7B-Level Models}}} \\
Qwen2-VL-7B~\citep{qwen2vl} & 2327 &54.1  & 16.3 & 31.9 & - \\
InternVL2-8B~\citep{internvl2} & 2210 &52.6  &18.4  &37.0  &1.9  \\
MiniCPM-Llama-V-2.5-8B~\citep{Minicpmv} 2024 & 45.8 &- & -&- &- \\
Cambrain-1-8B~\citep{cambrian}  &- & 42.7& - & -&- \\
LLava-Onevision-7B~\citep{Llava-onevision} &1998  &48.8  & - &26.2  & - \\
MiniCPM-V2.6~\citep{Minicpmv} & \textbf{2348} &49.8  &18.3 &25.7  & - \\
\rowcolor{gray!30}\textsc{OmniCaptioner} + Qwen2.5-7B-Instruct & 1824  & \textbf{54.5} & 26.4 &37.4 &\textbf{10.9}  \\
\rowcolor{gray!30}\textsc{OmniCaptioner} + DS-R1-Distill-Qwen-7B & 1942 & 47.5 & \textbf{36.2} & \textbf{40.5} & 7.8 \\
\cmidrule{1-6}
\multicolumn{6}{c}{\textit{\textbf{32B-Level Models}}} \\
InternVL-Chat-V1.5~\citep{internvl2} &2194 & 46.8 &  15.0  & 28.4 & 0.6 \\
InternVL2-26B~
\citep{internvl2}& 2260 & 51.2 &  17.0  & 31.1 & 3.5  \\
Cambrian-34B~\citep{cambrian}  & - & 49.7 & - & - & - \\
VILA-1.5-40B &-  & 55.1 & - & - & - \\
InternVL2-40B & \textbf{2307} & 55.2 & 16.9 &  36.3 & 3.9 \\
\rowcolor{gray!30}\textsc{OmniCaptioner} + Qwen2.5-32B-Instruct & 1831 & \textbf{59.7} &32.1  &39.7  &13.1 \\
\rowcolor{gray!30}\textsc{OmniCaptioner} + DS-R1-Distill-Qwen-32B& 2007 & 59.2 & \textbf{43.3} & \textbf{43.7} & \textbf{13.2 } \\
\cmidrule{1-6}
\multicolumn{6}{c}{\textit{\textbf{72B-Level Models}}} \\
Qwen2-VL-72B~\citep{qwen2vl} &\textbf{2482}  & 64.5 &25.9 &-  & 11.2 \\
InternVL2-76B~\citep{internvl2} &2414  & 62.7 &  23.6 & \textbf{42.8} & 5.5  \\
LLaVA-OneVision-72B~\citep{Llava-onevision} & 2261 & 56.8 & - & 39.1 & - \\

\rowcolor{gray!30}\textsc{OmniCaptioner} + DS-R1-Distill-Llama-70B & 2025 & \textbf{64.6} & \textbf{42.9} & 42.5 & \textbf{13.7} \\
\bottomrule[1pt]
\end{tabular}
}
\end{table}

\noindent \textbf{Efficient SFT Process.} The training paradigm of MLLMs typically consists of two sequential phases: pretraining on image-caption data, followed by Supervised Fine-Tuning (SFT). Empirical studies~\citep{chen2024compcap,jiang2025mme,mckinzie2024mm1} have demonstrated that diverse and high-quality image-caption data (\egno, composite images) can significantly enhance image-language alignment and subsequently promote performance on downstream tasks, such as Visual Question Answering (VQA). \textsc{OmniCaptioner} leverages diverse and high-quality domain data (\egno, table, chart, and so on) during the pretraining phase, enabling the model to acquire multi-domain knowledge. During the SFT phase, the multi-domain knowledge serves as a crucial foundation for rapid adaptation to downstream tasks across different domains.

\section{Experiment}

To evaluate \textsc{OmniCaptioner}, we conduct four primary experiments. The \textbf{first} experiment focuses on the caption evaluation from the perspective of objective metrics and subjective preference. The \textbf{second} experiment focuses on visual reasoning with a Caption-inserted Large Language Model. In this setup, detailed captions and corresponding questions are provided to the LLM, and its ability to answer the questions is evaluated. We use five benchmark datasets to assess the model's performance on this downstream task: MME~\citep{fu2023mme}, Mathverse~\citep{zhang2024mathverse}, Mathvision~\citep{mathvision}, MMMU~\citep{yue2024mmmu} and Olympiad bench~\citep{he2024olympiadbench}. For the LLMs, we select Qwen2.5-3B-Instruct~\citep{yang2024qwen2-5}, \tct{Qwen2.5-7B-Instruct~\citep{yang2024qwen2-5}, Qwen2.5-32B-Instruct~\citep{yang2024qwen2-5}}, DeepSeek-R1-Distill-Qwen-7B~\citep{Deepseek-r1}, DeepSeek-R1-Distill-Qwen-32B~\citep{Deepseek-r1}, and DeepSeek-R1-Distill-LLaMA-70B~\citep{Deepseek-r1}, all chosen for their strong reasoning capabilities. 
The \textbf{third} experiment involves finetuning the text-to-image generation model~\citep{lumina2,gao2024lumina,xie2025sana} such as SANA-1.0-1.6B~\citep{xie2025sana} with image-caption pairs generated by different captioners (\ieno, Qwen2-VL~\citep{qwen2vl}, \textsc{OmniCaptioner}). The training setting uses a resolution of 1024 $\times$ 1024. The model's generative performance is then evaluated on the GenEval~\citep{ghosh2023geneval}.
The \textbf{fourth} experiment evaluates the efficiency of the SFT process. For this, we select the LLaVA-OneVision~\citep{Llava-onevision} data from the OV stage with chain-of-thought enhancement to assess the SFT version of \textsc{OmniCaptioner} across multiple commonly-used benchmarks~\citep{fu2023mme,yue2024mmmu,masry2022chartqa,mathew2021docvqa,mathvision,zhang2024mathverse,lu2023mathvista}.

\subsection{Main Results}

\begin{table*}[t]
\centering
\begin{minipage}[t]{0.45\textwidth}
\centering
\caption{Caption Metrics comparison across models. \textsc{OmniCaptioner} achieves the highest score in all metrics.}
\label{captoin_metric}
\begin{adjustbox}{width=\linewidth}
\begin{tabular}{lccc}
\toprule
\textbf{Metric} & LLaVA-OV-7B & Qwen2-VL-7B & OmniCaptioner \\
\midrule
BLEU & 14.18 & 21.70 & \textbf{22.35} \\
CLIPScore & 30.12 & 32.71 & \textbf{34.05} \\
CAPTURE & 62.40 & 64.38 & \textbf{64.88} \\
\bottomrule
\end{tabular}
\end{adjustbox}
\label{tab:caption_metrics}
\end{minipage}
\hfill
\begin{minipage}[t]{0.48\textwidth}
\centering
\caption{User study results on caption preference (\%). \textsc{OmniCaptioner} is preferred by human evaluators across both image types.}
\label{captoin_usr}
\begin{adjustbox}{width=\linewidth}
\begin{tabular}{lcc}
\toprule
\textbf{Domain} & Qwen2-VL-7B & \textsc{OmniCaptioner} \\
\midrule
Non-Natural Images & 43.3 & \textbf{56.7} \\
Natural Images & 48.8 & \textbf{51.2} \\
\bottomrule
\end{tabular}
\end{adjustbox}
\label{tab:user_study}
\end{minipage}
\end{table*}

\noindent \textbf{Caption Quality Comparison.}  To evaluate the quality of generated captions, we conduct a comprehensive comparison using both objective metrics and subjective human evaluation. For the objective evaluation, we measure BLEU~\citep{papineni2002bleu}, CLIPScore~\citep{hessel2021clipscore}, and CAPTURE scores~\citep{capture}. As shown in Table~\ref{captoin_metric}, our method, \textsc{OmniCaptioner}, consistently outperforms strong baselines such as LLaVA-OneVision and Qwen2-VL-7B-Instruct, achieving the highest scores across all metrics: 22.35 on BLEU, 34.05 on CLIPScore, and 64.88 on CAPTURE, demonstrating superior alignment with both textual references and visual semantics. In addition to automated evaluation, we conduct a user study to assess human preference. We collect 90 images from three diverse sources—MMMU, ChartQA, and MME—and categorize them into natural and non-natural domains. Captions are generated by both \textsc{OmniCaptioner} and Qwen2-VL-7B-Instructand evaluated in a blind pairwise comparison setting by 10 human raters. Each rater selects the preferred caption based on relevance, informativeness, and fluency. As shown in Table~\ref{captoin_usr}, \textsc{OmniCaptioner} is preferred in 56.7\% of non-natural image cases and 51.2\% of natural image cases, indicating a consistent advantage in human-perceived quality across different image types.

\noindent \textbf{Improved Visual Reasoning with LLMs.} Our experimental results of Table~\ref{caption_LLM} demonstrate that integrating captions into reasoning-enhanced Large Language Models (LLMs), without any additional fine-tuning, achieves state-of-the-art performance across multiple reasoning benchmarks, including MathVision~\citep{mathvision}, MathVerse~\citep{zhang2024mathverse}, MMMU~\citep{yue2024mmmu}, and Olympiad bench~\citep{he2024olympiadbench}. This highlights the power of \textsc{OmniCaptioner} in boosting reasoning capabilities for multiple visual tasks.
Specifically, \textsc{OmniCaptioner}-inserted LLMs significantly outperform existing models in MathVision across multiple model sizes, underscoring the enhancement of reasoning ability for complex visual and mathematical tasks. Notably, \tct{\textit{OmniCaptioner + DS-R1-Distill-Qwen-7B} and \textit{OmniCaptioner + DS-Distill-Qwen-32B}} demonstrate exceptional performance on MathVerse benchmark, significantly outperforming previous models. These results further validate the efficacy of caption-based pretraining in bridging the LLM's comprehension of visual geometry content.
In the MMMU benchmark, \tct{\textit{OmniCaptioner + DS-R1-Distill-Qwen-72B}} approaches the performance of Qwen2-VL-72B, with a minimal gap between them. This result serves as strong evidence that caption integration with reasoning-enhanced LLMs leads to significant visual understanding and reasoning for multidisciplinary content. 

The successful integration of captions with LLMs across scales, from 3B to 72B, underscores that \tct{\textsc{OmniCaptioner}} consistently enhances LLMs’ reasoning abilities for visual tasks, yielding improvements irrespective of model size. These results highlight that our unified pretraining methodology, leveraging large-scale caption data, is a highly effective strategy for advancing visual reasoning across diverse tasks, outperforming existing approaches even when compared to large-scale fine-tuning methods.

\noindent \textbf{Enhanced Image Generation.} 
As illustrated in Tab.~\ref{tab:GenEval}, to validate the importance of caption accuracy in T2I generation, our model demonstrates significant performance improvements over the Qwen2-VL-Instruct ~\citep{qwen2vl} caption and original SANA, on GenEval benchmark. 
The original SANA model achieves a 64.61 overall score on GenEval, which is significantly improved to 65.27 with Qwen2-VL-Instruct and further to 67.58 with\textsc{OmniCaptioner}. This +2.97 absolute gain over the vanilla SANA model underscores the effectiveness of high-quality captions in guiding T2I generation.
Also, our \textsc{OmniCaptioner} outperforms Qwen2-VL-Instruct across various aspects (except colors), showcasing the enhanced accuracy of our caption generation.

\begin{table}
\centering
\caption{Performance comparison of models trained with different captioners on GenEval~\citep{ghosh2023geneval} (Resolution: 1024 $\times$ 1024).}
\label{tab:GenEval}
\renewcommand{\arraystretch}{1.0}
\resizebox{0.88\linewidth}{!}{\begin{tabular}{l  c c c c c c }
\toprule[1pt]
\multirow{2}{*}{\textbf{Methods}} & \multicolumn{6}{c}{\textbf{GenEval $\uparrow$}} \\
\cmidrule(lr){2-7}

 & Color Attri. & Sin. Obj. & Pos. & Colors & Counting & Overall  \\
\cmidrule(lr){1-7}
SANA-1.0-1.6B~\citep{xie2025sana}  & 38.50 & 98.75 & 21.25  & 86.70 & \textbf{65.31} & 64.61  \\
SANA-1.0-1.6B + Qwen2-VL~\citep{qwen2vl}  & 44.29 & 98.44 & 26.64  & \textbf{86.97} & 57.81 &  65.27  \\
\rowcolor{gray!30} SANA-1.0-1.6B + \textsc{OmniCaptioner}  & \textbf{46.00} & \textbf{99.06} & \textbf{29.50}  & 84.57 & 64.06 & \textbf{67.58} \\
\bottomrule[1pt]
\end{tabular}}
\end{table}

\noindent \textbf{Efficient SFT.} In Tab.~\ref{tab:SFT_1}, we compare the performance of several models on visual perception and reasoning tasks, including \textit{LLaVA-OV-7B(SI)}, \textit{LLaVA-OV-7B}, \textit{Qwen2-VL-base+OV SFT}, and our proposed \textit{OmniCaptioner+OV SFT} model. While \textit{LLaVA-OV-7B (SI)} and \textit{LLaVA-OV-7B} use significantly larger datasets for SFT – 3.2M and 4.8M examples, respectively – our \textit{OmniCaptioner+OV SFT} achieves comparable results with just 1.6M SFT examples used during the one-vision (OV) stage. 
A key difference lies in the unified pretraining phase of \textsc{OmniCaptioner}, which utilizes a diverse caption-based dataset prior to the SFT stage. This step equips the model with richer domain knowledge, enabling it to excel in visual instruction-following tasks despite fewer SFT examples. It also reveals that \textit{Qwen2-VL-base + SFT} lags behind \textit{OmniCaptioner + OV SFT}, indicating \textsc{OmniCaptioner}'s superior visual perception capabilities.

\begin{table}[t]
\centering

\caption{SFT performance comparison across diverse evaluation benchmarks. OmniCaptioner + OV SFT denotes the SFT model based on \textsc{OmniCaptioner}, while Qwen2-VL-base + OV SFT is based on Qwen2-VL-Base. LLaVA-OV-7B (SI) represents the model after the single-image training in LLaVA-OneVision~\citep{Llava-onevision}.}
\label{tab:SFT_1}
\small
\renewcommand{\arraystretch}{1.0}
\resizebox{1\linewidth}{!}{
\begin{tabular}{l c c c c  c c c }
\toprule[1pt]
\textbf{SFT Model} & \textbf{Data} &\textbf{MME} & \textbf{MMMU}    & \textbf{MathVerse} & \textbf{MathVista} & \textbf{DocVQA} & \textbf{ChartQA} \\

\cmidrule(lr){1-2} \cmidrule(lr){3-4} \cmidrule(lr){5-6} \cmidrule{7-8}

LLaVA-OV-7B (SI)~\citep{Llava-onevision} & 3.2M &2109    & 47.3 & 26.9 & 56.1 & 89.3/86.9 & 78.8 \\
LLaVA-OV-7B~\citep{Llava-onevision} &4.8M & 1998   & 48.8 & 26.2 & 63.2 & 90.2/87.5 & 80.0 \\
Qwen2-VL-Base+OV SFT &1.6M &1905  & 44.4   & 24.9 & 53.8 & 84.2/- & 53.5 \\
\rowcolor{gray!30}\textsc{OmniCaptioner}+OV SFT &1.6M &2045  & 46.6   &25.8  & 57.4 & 91.2/- & 79.0 \\
\bottomrule[1pt] 
\end{tabular}
}
\end{table}

\begin{table}[t]
\centering
\caption{Comparing different captioners through experiments with captioner-inserted LLM on multiple visual benchmarks.}
\label{tab:different_captioner}
\small
\renewcommand{\arraystretch}{1.0}
\resizebox{1\linewidth}{!}{\begin{tabular}{l l c c c c }
\toprule[1pt]
\textbf{Captioner Selection} & \textbf{LLM} & \textbf{MME} & \textbf{MMMU} & \textbf{MathVision} & \textbf{MathVerse} \\
\cmidrule(lr){1-6}
llava-onevision-qwen2-7b-ov & DS-R1-Distill-Qwen-7B  &1646 & 22.4 &31.7 & 36.6 \\
InternVL2-8B & DS-R1-Distill-Qwen-7B  & 1789 &23.1  & 34.4&  39.9\\
Qwen2-VL-7B-Instruct & DS-R1-Distill-Qwen-7B  &1914  &42.4  & 31.6 & 33.0   \\
\rowcolor{gray!30}\textsc{OmniCaptioner} (ours) & DS-R1-Distill-Qwen-7B  & \textbf{1942} & \textbf{47.5} & \textbf{36.2} & \textbf{40.5}  \\

\bottomrule[1pt]
\end{tabular}}
\end{table}

\begin{figure}[tbp]
  \centering
   \includegraphics[width=0.8\linewidth]{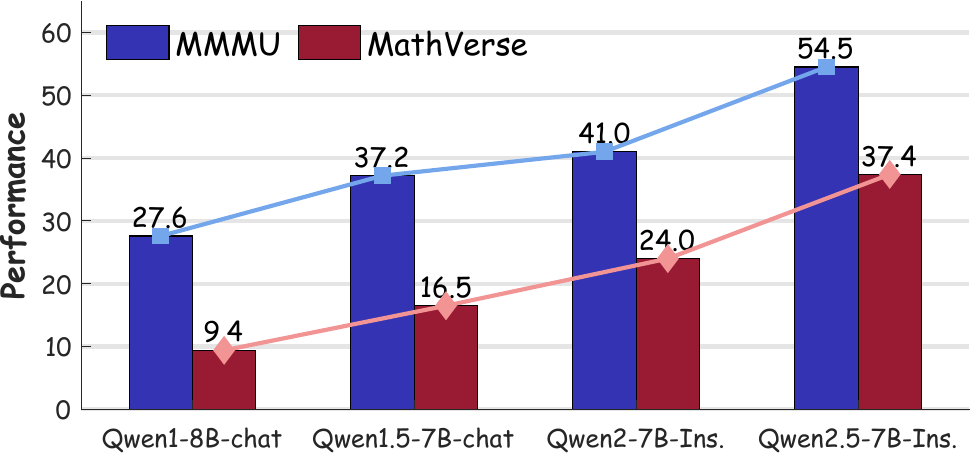}
   \caption{Integrate \textsc{OmniCaptioner} into different versions of LLMs, enabling them to handle tasks in multimodal scenarios.}
   \label{fig:qwenlm-version}
\end{figure}

\subsection{Discussions and Findings}
We consider conducting three important experiments when combining \textsc{OmniCaptioner} with reasoning-enhanced LLMs. First, we evaluate effectiveness using different Qwen versions. Second, we aim to explore the extent to which Qwen2-VL-Instruct (without image input) and mainstream reasoning-enhanced LLMs rely on visual modality information to solve visual reasoning tasks. Third, we compare \textsc{OmniCaptioner} to Qwen2-VL-Instruct by modifying the captions provided to the reasoning-enhanced LLMs. For more visualization results of image captioning, video captioning, and text-to-image generation task, please refer to Appendix~\ref{sec:cap_vis} and Appendix~\ref{sec:text-to-img}.

\noindent \textbf{Effect of Different Qwen-Family Versions.}
Fig.~\ref{fig:qwenlm-version} illustrates the performance progression of combining \textsc{OmniCaptioner} with different versions of Qwen on MMMU and MathVerse. As Qwen evolves from Qwen1-8B-chat to Qwen2.5-7B-Instruct, there is a steady improvement in visual reasoning capabilities, driven by the pixel-to-word captioning ability of \textsc{OmniCaptioner}.
As illustrated in Fig.~\ref{fig:wovisualinput}, the performance comparison between \textit{OmniCaptioner + Qwen2.5-7B-Base}, \textit{OmniCaptioner + Qwen2.5-7B-Instruct} and \textit{OmniCaptioner + DS-R1-Distill-Qwen-7B} highlights the advantage of integrating the DeepSeek Distilled Qwen2.5, which excels in mathematical reasoning. 
The distilled variant (\textit{DS-R1-Distill-Qwen-7B}) achieves the highest accuracy across \textit{\textbf{MME (1942)}}, \textit{\textbf{MathVision (36.2)}}, and \textit{\textbf{MathVerse (40.5)}}, emphasizing the benefits of distilled reasoning ability. 
In contrast, \textit{Qwen2.5-7B-Instruct} is better suited for general world knowledge tasks, as reflected in its improved performance on the \textit{\textbf{MMMU (54.5)}}.

\noindent \textbf{Impact of Visual Modality on Reasoning-Enhanced LLMs.} 
From Fig.~\ref{fig:wovisualinput}, the performance of \textit{Qwen2-VL-7B (NA)} and \textit{DeepSeek-Distill-Qwen-7B} suggests that the absence of image input significantly restricts their ability to solve visual reasoning tasks.  
In contrast, \textit{OmniCaptioner + DS-R1-Distill-Qwen-7B}, which retains visual modality, achieves substantially higher accuracy than its non-visual input LLM, highlighting the critical role of visual information in enhancing reasoning capabilities.  
Furthermore, non-visual input LLM \textit{DS-R1-Distilled-Qwen-7B} significantly outperforms the no-image MLLM (\ieno, \textit{Qwen2-VL-Instruct-7B}) on MathVision and MathVerse, demonstrating the superior reasoning ability of R1 Serious model.

\noindent \textbf{Effect of Different Captioners.} Tab.~\ref{tab:different_captioner} presents a comparative analysis of different captioners on multiple perception and reasoning benchmarks. Our model, incorporating DeepSeek-Distill-Qwen2.5-7B, achieves superior performance across all evaluated tasks, significantly outperforming previous approaches.
These results highlight the effectiveness of \textsc{OmniCaptioner}, whose captions provide more precise and contextually accurate descriptions than those generated by Qwen2-VL-7B-Instruct. The enhanced caption quality contributes to improved visual reasoning tasks, particularly in tasks requiring multi-step inference and detailed visual understanding.

\section{Conclusion}

We have introduced \textsc{OmniCaptioner}, a unified framework that bridges visual and textual modalities through fine-grained pixel-to-text mapping across diverse domains, including natural images, visual-text images and structured images. By converting low-level visual patterns into semantically rich captions, our approach empowers reasoning-enhanced LLMs (\egno, DeepSeek-R1) to achieve enhanced visual reasoning, and enables precise text-to-image generation through comprehensive semantic preservation. This work pioneers a scalable paradigm for multimodal alignment and reasoning, achieving seamless visual-language interoperability without costly label-supervised fine-tuning.

\section*{Acknowledgement}
The research was supported by Shanghai Artificial Intelligence Laboratory, a locally commissioned task from the Shanghai Municipal Government, the Shanghai Municipal Science and Technology Major Project, and Shanghai Rising Star Program (Grant No. 23QD1401000).

\bibliography{main}
\bibliographystyle{colm2025_conference}

\appendix

\section{OmniCaptioner Dataset Composition}
\label{sec:data_comp}

\begin{figure}[htbp]
  \centering
   \includegraphics[width=0.70\linewidth]{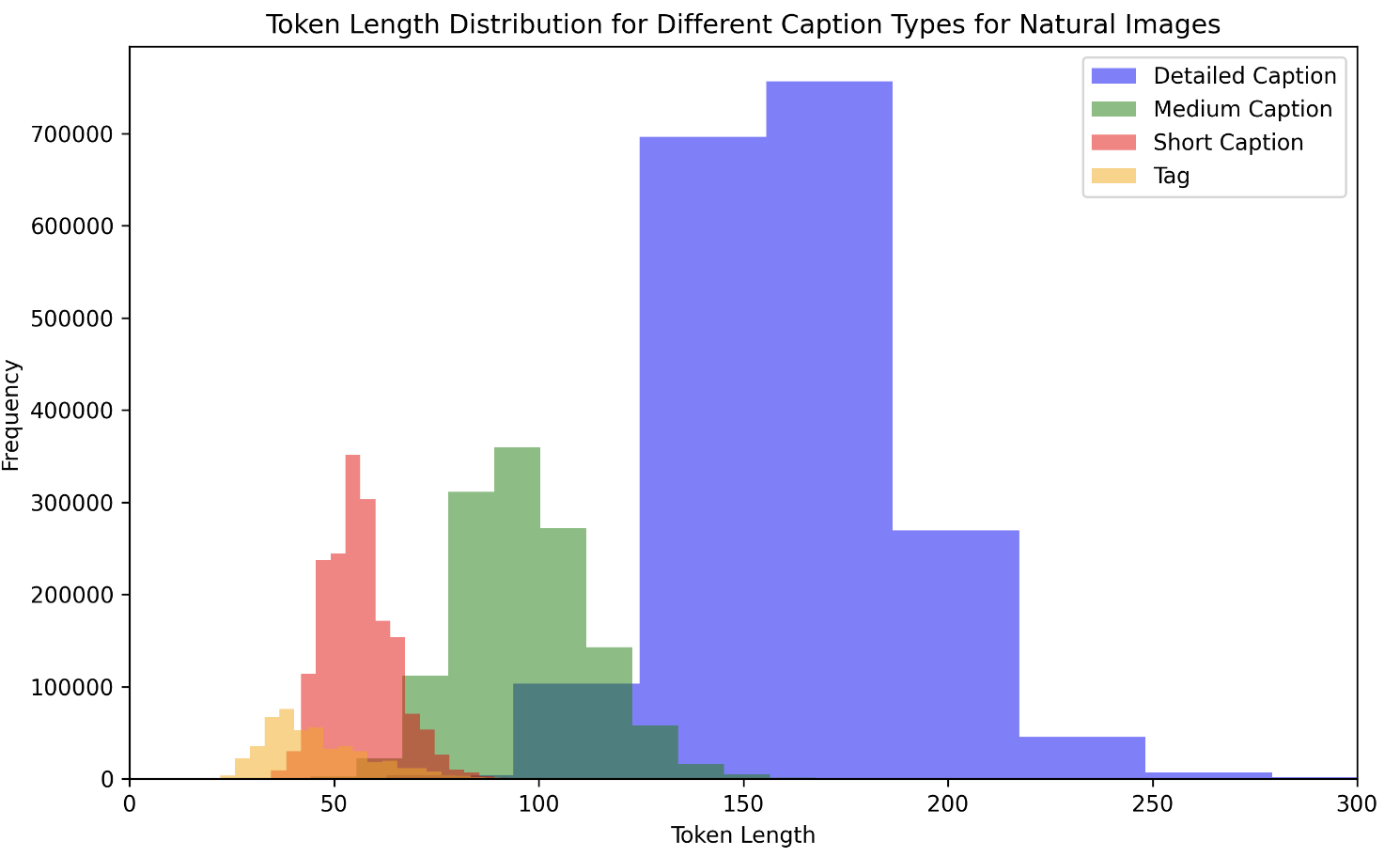}
    \vspace{-0.2cm}
   \caption{Token length distribution for natural images.}
   \label{fig:token_len_dis}
\end{figure}

\begin{figure}[h]
  \centering
   \includegraphics[width=0.7\linewidth]{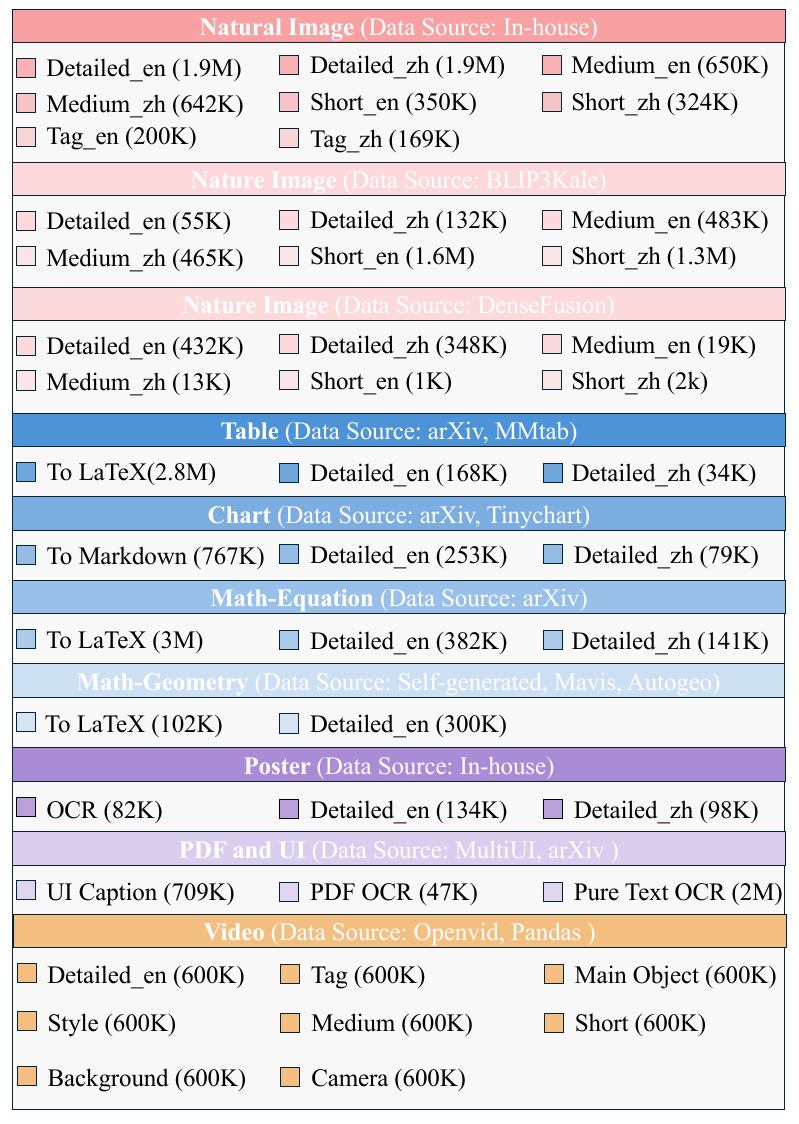}
\vspace{-0.2cm}
   \caption{Dataset composition for pretraining \textsc{OmniCaptioner}.}
   \label{fig:data_src}
\end{figure}

As shown in Fig.~\ref{fig:data_src}, the \textsc{OmniCaptioner} dataset is a large-scale multimodal benchmark comprising images, tables, charts, mathematical geometry/equations, posters, PDFs, UI elements, and videos, with captions available in both English and Chinese. The dataset includes natural images sourced from in-house collections, BLIP3Kale~\citep{awadalla2024blip3}, and DenseFusion~\citep{li2024densefusion}. Tabular data are collected from the arXiv website and the open-source MMTab dataset \citep{zheng2024multimodal}, while chart data originate from arXiv website and TinyChart \citep{zhang2024tinychart}. Mathematical content, including equations and geometric structures, is sourced from arXiv and generated from datasets such as MAVIS \citep{zhang2024mavis} and AutoGeo \citep{huang2024autogeo}. UI data are obtained from the MultiUI dataset \citep{liu2024harnessing}, while poster images feature OCR-based captions. Video captions are derived from OpenVid \citep{nan2024openvid} and Panda \citep{chen2024panda}, covering multiple attributes such as detailed descriptions, style, background, tags, camera angles, and object information. 
Fig.~\ref{fig:token_len_dis} illustrates the token length distribution for different caption types associated with natural images, categorized into detailed, medium, short, and tag captions. 


\section{Experimental Setup}

We fine-tune the Qwen2-VL-7B-Instruct model on a large-scale captioning dataset using 64 A100 GPUs. The training process is distributed using torchrun with the DeepSpeed ZeRO-3 optimization strategy.

\noindent \textbf{Hyperparameters:}
\begin{itemize}
    \item \textbf{Batch Size:} 256 (1 per device, with gradient accumulation of 8)
    \item \textbf{Learning Rate:} 1e-5 (base model), 1e-5 (merger module), 2e-6 (vision tower)
    \item \textbf{Weight Decay:} 0.0
    \item \textbf{Warmup Ratio:} 3\%
    \item \textbf{Scheduler:} Cosine decay
    \item \textbf{Precision:} BF16 enabled,
    \item \textbf{Gradient Checkpointing:} Enabled
\end{itemize}

\noindent \textbf{Training Details:}
\begin{itemize}
    \item \textbf{Image Resolution:} From \textbf{2$\times$28$\times$28} to \textbf{6400$\times$28$\times$28} pixels
    \item \textbf{Epochs:} 1
\end{itemize}


\section{Prompt for Caption Annotation}

\begin{tcolorbox}[
  colback=white,
  colframe=black,
  coltitle=white,
  title=Natural Image Annotation Prompt for GPT-4o,
  fonttitle=\bfseries,
  colbacktitle=black,
  arc=1mm,
  boxrule=0.8pt,
  width=\textwidth,
  breakable
]

You are an expert in image captioning, segmentation labeling, and stylistic descriptions at the level of an Oscar-winning cinematographer, photographer, or illustrator. Your task is to give me an extremely information-dense description of each image I send you. 

Remember that you may need to caption images from all visual domains imaginable: Photography, Movie Stills, Animated Pixar movies, Sketches, IKEA assembly instruction diagrams, Screenshots, UIs, Cave Paintings, Abstract Art, Product Photography, all forms of illustrations, and many more genres. Therefore, you need to be quite descriptive and effective at identifying the artistic medium and render technique utilized. Avoid unnecessary repetitions and redundancy. Only write what you feel reasonably confident about. Occasional mistakes are okay, but do not hallucinate what you do not actually see in each image.

\textbf{Your response should for the most part answer three questions:}
\begin{enumerate}[label=\arabic*.]
  \item How would you describe this image and its environment overall? \\
  (e.g., "Photo portrait of a white, middle aged man in front of a white background looking to the right.")
  \item What are all objects you see in this image and where exactly are they placed? \\
  (e.g., "A yellow Taxi driving forwards in the left foreground. Pedestrian crossing and cracked asphalt street in the center. Many cars in the background. New York buildings skyline in the background.")
  \item What are all purely stylistic properties that this image shows? \\
  (e.g., "Underexposed, dark, moody, Photorealistic, Shallow depth of field, natural lighting, golden hour, warm color palette, high contrast, automotive photography, tack sharp, glossy texture, muted brown earth tones, low angle perspective, rustic urban landscape.")
\end{enumerate}

\textbf{Style and Formatting Instructions:}
\begin{itemize}
  \item Fuse all answers into a single, coherent string.
  \item Do not use semantic labels such as “Stylistic properties:”.
  \item Use full sentences for overall descriptions, and tightly punctuated keywords for visual style.
  \item Do not start with “The image depicts” or “This image has”. Go straight to the content.
  \item Avoid phrases like “suggesting that”, “potentially”, “might be”. Be visually confident.
  \item Describe all identifiable objects: position, size, color, material, orientation, relation to others.
  \item Describe stylistic traits: lighting, color grading, rendering method, medium, realism level.
  \item Format output as a single dense caption. Use periods or commas, but no lists or line breaks.
\end{itemize}

\textbf{Output Format:} One highly descriptive string. No list. No section labels. No bullets. No line breaks.

\textbf{Examples:}

\textit{Example 1:} Photo of a sleek, grey Ferrari F8 parked in a narrow cobblestone alleyway of an old Italian village. The car is positioned in the center of the frame, facing slightly to the right. The background features rustic buildings with weathered, beige plaster walls and wooden shutters. A building to the left has a wooden door and an arched entrance partially covered by ivy. A vibrant red bougainvillea climbs up the left building, and a green bush with yellow flowers is visible next to it. In the distant background, there's a hillside with dense greenery. The foreground includes out-of-focus branches with yellow leaves, framing the image. Photo, underexposed, dark mood, medium depth of field, soft natural lighting, golden hour, warm color palette, photorealistic, high contrast, automotive photography, tack sharp, glossy texture, nostalgic, serene, visually balanced.

\textit{Example 2:} Ethereal 3D image of a character playing on a piano suspended in a blue dreamlike environment. Scene from Pixar's animated movie "Soul". The character, a man wearing a hat and glasses, is seated on a stool to the left of the piano. The piano is a glossy black grand piano, centrally positioned, with its lid open. The scene is bathed in vibrant purple and blue lighting, creating an ethereal and otherworldly atmosphere. The background is filled with abstract light patterns and gradients, enhancing the surreal feel of the image. Pixar animation, medium depth of field, soft diffused lighting, neon color palette, photorealistic textures, high contrast, ethereal, whimsical, visually balanced, dynamic composition, dramatic lighting effects, digital animation.

\end{tcolorbox}

\begin{tcolorbox}[
  colback=white,
  colframe=black,
  coltitle=white,
  title=Video Annotation Prompt for GPT-4o,
  fonttitle=\bfseries,
  colbacktitle=black,
  arc=1mm,
  boxrule=0.8pt,
  width=\textwidth,
  breakable
]

You are describing a video represented by frames extracted at a rate of one frame per second. Based on these frames, provide detailed captions in English from the following aspects:

\textbf{1. Short Caption:} Summarize the video in one detailed sentence, capturing key actions and the overall mood.

\textbf{2. Background Caption:} Describe the background, including objects, location, weather, time, and dynamic elements like movements.

\textbf{3. Main Object Caption:} Describe the main subject's actions, attributes, interactions, and movements across the frames, including changes in posture, expression, or speed.

\textbf{4. Reference Caption:} Provide a detailed, dense caption (around 250 words) describing all visible actions, environmental details, and emotional atmosphere. Use a structured approach covering:
\begin{itemize}
    \item Subject
    \item Subject actions
    \item Environment and background
    \item Visual language (style, composition, lighting)
    \item Camera language (movement, angles, focal length, shot sizes)
\end{itemize}
Highlight the mood and tone, and create a vivid narrative rich enough for AI to recreate the video.

\textbf{5. Standard Summary:} Provide a concise, approximately 100-word summary that highlights the main actions, key subjects, and important environmental details.

\textbf{6. Style Tags:} Provide a single, comma-separated string of tags (at least 5) that includes video types, video style, and any relevant attributes.

\textbf{7. Key Tags:} Provide a single, comma-separated string of tags including key objects, people, or entities (3–5), location, time, environment (2–3), notable qualities (2–4), video style, and camera techniques (1–2).

\vspace{1em}
\textbf{Important Camera Work Requirement:}

For all sections, descriptions related to the camera work (including shot types, camera angles, and camera movements) should primarily reference the following camera caption provided by the user:

\texttt{\{camera\_caption\_only\}}

This includes all mentions of how the video is framed, how the camera moves, and any stylistic elements related to the camera language. If the provided camera caption contains significant errors or inconsistencies, you may adapt the descriptions as needed, ensuring they remain accurate and cohesive. Avoid forcing unnecessary mentions of camera techniques.

\vspace{1em}
\textbf{Important Guidelines:}
\begin{itemize}
    \item Avoid describing each frame individually or using phrases like “first frame”.
    \item Do not start with “The scene...”, “In this video...”, etc. Write in vivid, flowing narrative form.
    \item For \textit{Reference Caption} and \textit{Standard Summary}, avoid any reference to "the video".
    \item Be cohesive and immersive. Avoid short descriptive fragments; instead use continuous, vivid narration.
    \item Strictly follow the format with all 7 sections labeled.
    \item Never decline a description. If objects or individuals are unidentifiable, describe their visual features or behavior.
    \item Only include camera work details if they align with the provided camera caption or visual evidence.
\end{itemize}

\end{tcolorbox}

\begin{tcolorbox}[
  colback=white,
  colframe=black,
  coltitle=white,
  title=Poster Annotation Prompt for GPT-4o,
  fonttitle=\bfseries,
  colbacktitle=black,
  arc=1mm,
  boxrule=0.8pt,
  width=\textwidth,
  breakable
]

You are an AI assistant specialized in analyzing Poster images and converting them into a structured markdown format.
You need to provide a detailed caption for an English poster. The main content of the poster includes text and non-text elements. Based on these elements, provide a concise English poster description in the order in which they appear in the poster, and describe them according to the following requirements:

\textbf{Instructions:}
\begin{enumerate}[label=\textbf{\arabic*.}, leftmargin=1.2em]
    \item Describe the textual and visual elements in the order they appear in the poster:
    
    \begin{itemize}
        \item For textual content: specify the font type (e.g., Heiti, Songti, Kaiti, Dengxian, handwriting), font color, font size, position (e.g., top, bottom, top-left, bottom-right, center), alignment (e.g., centered, left-aligned, right-aligned), whether it is obstructed, and layout characteristics (e.g., vertical, horizontal).
    \end{itemize}

    \begin{itemize}
        \item For visual elements: describe their properties (e.g., color, shape, size, dynamic or static, texture), position (e.g., centered, dispersed), layering with respect to text or other elements, and any decorative effects (e.g., border, shadow, gradient, texture).
    \end{itemize}
    \item Describe the layout and interaction between text and visual elements, including their spatial relationships (e.g., overlap, separation, symmetry).
    \item Provide an overall assessment of the poster’s style (e.g., bright, minimalistic, vintage, modern, tech-oriented, natural, artistic).
    \item Avoid speculating on the poster’s topic, narrative, or intention—focus solely on visual and structural features.
    \item Keep the description concise and accurate, focusing on visual aspects. Do not include unrelated content.
\end{enumerate}

\end{tcolorbox}


\begin{tcolorbox}[
  colback=white,
  colframe=black,
  coltitle=white,
  title=GUI Annontation Prompt for GPT-4o,
  fonttitle=\bfseries,
  colbacktitle=black,
  arc=1mm,
  boxrule=0.8pt,
  width=\textwidth,
  breakable
]

You are an AI assistant specialized in analyzing Graphical User Interface (GUI) images and converting them into structured markdown format. GUI images often contain background, navigation, interaction, visual and text information, layout, and icons. Your task is to: 

Provide detailed annotation of the Graphical User Interface (GUI) image. Based on the GUI's visual, text elements and layout, provide detailed descriptions in the following aspects:

1. \textbf{Brief description}: Summarize the GUI's main purpose and content in one concise but specific sentence.

2. \textbf{Detailed extraction}:
    \begin{itemize}
        \item If the GUI image contains background elements, describe background elements (e.g., colors, images, dynamic elements, and so on).
        \item Extract all elements from right to left and from top to bottom.
        \item Extract the content of the GUI image in detail and completely, without missing any part.
        \item Don’t miss any text that appears on the image.
        \item Use markdown format.
    \end{itemize}

3. \textbf{Description of interactive elements}: If the GUI image includes interactive elements (e.g., search boxes, buttons, and so on), describe them and their functionality and usage.

4. \textbf{Overall description}: Provide a summary of about 100 words, summarizing the main functions and usage scenarios of the GUI display page.

\textbf{Instructions:}
\begin{itemize}
    \item Please structure your response as follows: 1. Brief description, 2. Detail extraction, 3. Description of interactive elements, 4. Overall description.
    \item Ensure that the layout and visual language mentioned in all sections are consistent with this description.
    \item If you find major errors or inconsistencies in the description, you can adjust it as needed, but you must ensure accuracy and consistency.
    \item Please provide content strictly in the specified format, ensuring that all 4 sections are covered.
    \item Do not refuse any description request, even if the specific content cannot be identified, describe elements of the GUI image by inferring the characteristics.
    \item Ensure that the text description and visual language are consistent, but do not over-emphasize certain details or repeat content.
\end{itemize}
\end{tcolorbox}

\vspace{1cm}


\begin{tcolorbox}[
    enhanced,
    breakable,
    colback=white,
    colframe=black,
    coltitle=white,
    colbacktitle=black,
    title=Caption Extension Prompt for Medium Caption Using Qwen2.5-32B-Instruct,
    fonttitle=\bfseries,
    width=\textwidth,
    boxrule=0.7pt,
    arc=2mm,
    top=6pt,
    bottom=6pt,
    left=6pt,
    right=6pt
]

\textbf{Task:} Summarize the following long caption into a medium-length caption.

The medium caption should be shorter than the long caption. It should retain the key information from the long caption while improving clarity and brevity.

\textbf{Input:} Long Caption: \{caption\}

\end{tcolorbox}

\begin{tcolorbox}[
  colback=white,
  colframe=black,
  coltitle=white,
  title=Caption Extension Prompt for Short Caption Using Qwen2.5-32B-Instruct,
  fonttitle=\bfseries,
  colbacktitle=black,
  arc=1mm,
  boxrule=0.8pt,
  width=\textwidth,
  breakable
]

\textbf{Task:} Summarize the following medium caption into a short-length caption.

The short caption should be shorter than the long caption. It should retain the key information from the short caption while improving clarity and brevity.

\textbf{Input:} Medium Caption: \{caption\}

\end{tcolorbox}

\begin{tcolorbox}[
    enhanced,
    breakable,
    colback=white,
    colframe=black,
    coltitle=white,
    colbacktitle=black,
    title=Caption Extension Prompt for Key Tags Using Qwen2.5-32B-Instruct,
    fonttitle=\bfseries,
    width=\textwidth,
    boxrule=0.7pt,
    arc=2mm,
    top=6pt,
    bottom=6pt,
    left=6pt,
    right=6pt
]

You are given a detailed caption in English. Your task is to extract key tags (keywords) from the caption that capture the main concepts or themes.

Summarize the key tags. Each set of tags should be concise and represent the core ideas of the caption.

Use the following JSON format for your output:

\texttt{\{"tag1, tag2, tag3,..."\}}

The provided caption: \{caption\}

\end{tcolorbox}

\begin{tcolorbox}[
    enhanced,
    breakable,
    colback=white,
    colframe=black,
    coltitle=white, 
    colbacktitle=black,
    title=Caption Translation Prompt Using Qwen2.5-32B-Instruct (English to Chinese),
    fonttitle=\bfseries,
    width=\textwidth,
    boxrule=0.7pt,
    arc=2mm,
    top=6pt,
    bottom=6pt,
    left=6pt,
    right=6pt
]

You are given a detailed caption in English. Your task is to translate this detailed English caption to a Chinese caption that preserves the meaning and visual richness of the original.

The provided caption: \{caption\}

\end{tcolorbox}

\begin{tcolorbox}[
    enhanced,
    breakable,
    colback=white,
    colframe=black,
    coltitle=white, 
    colbacktitle=black,
    title=Caption Extension Prompt for Detailed Analysis of Table (Qwen2-VL-76B-Instruct),
    fonttitle=\bfseries,
    width=\textwidth,
    boxrule=0.7pt,
    arc=2mm,
    top=6pt,
    bottom=6pt,
    left=6pt,
    right=6pt
]
Please help me analyze the table image and the corresponding LaTeX code. The provided LaTeX code represents the structure of the table.

Provided LaTeX Code: \{Latex\}

\end{tcolorbox}

\begin{tcolorbox}[
    enhanced,
    breakable,
    colback=white,
    colframe=black,
    coltitle=white, 
    colbacktitle=black,
    title=Caption Extension Prompt for Detailed Analysis of Equation (Qwen2-VL-76B-Instruct),
    fonttitle=\bfseries,
    width=\textwidth,
    boxrule=0.7pt,
    arc=2mm,
    top=6pt,
    bottom=6pt,
    left=6pt,
    right=6pt
]
Please help me analyze the equation image and the corresponding LaTeX code. The provided LaTeX code represents the structure of the equation.

Provided LaTeX Code: \{Latex\}

\end{tcolorbox}

\begin{tcolorbox}[
    enhanced,
    breakable,
    colback=white,
    colframe=black,
    coltitle=white, 
    colbacktitle=black,
    title=Caption Extension Prompt for Detailed Analysis of Chart (Qwen2-VL-76B-Instruct),
    fonttitle=\bfseries,
    width=\textwidth,
    boxrule=0.7pt,
    arc=2mm,
    top=6pt,
    bottom=6pt,
    left=6pt,
    right=6pt
]
Please help me analyze this image of chart and corresponding markdown in detail.

Provided markdown
format of this chart image: \{markdown\}

\end{tcolorbox}

\section{System Prompt Example for \textsc{OmniCaptioner}}

\begin{figure*}[htbp]
  \centering
   \includegraphics[width=1.0\linewidth]{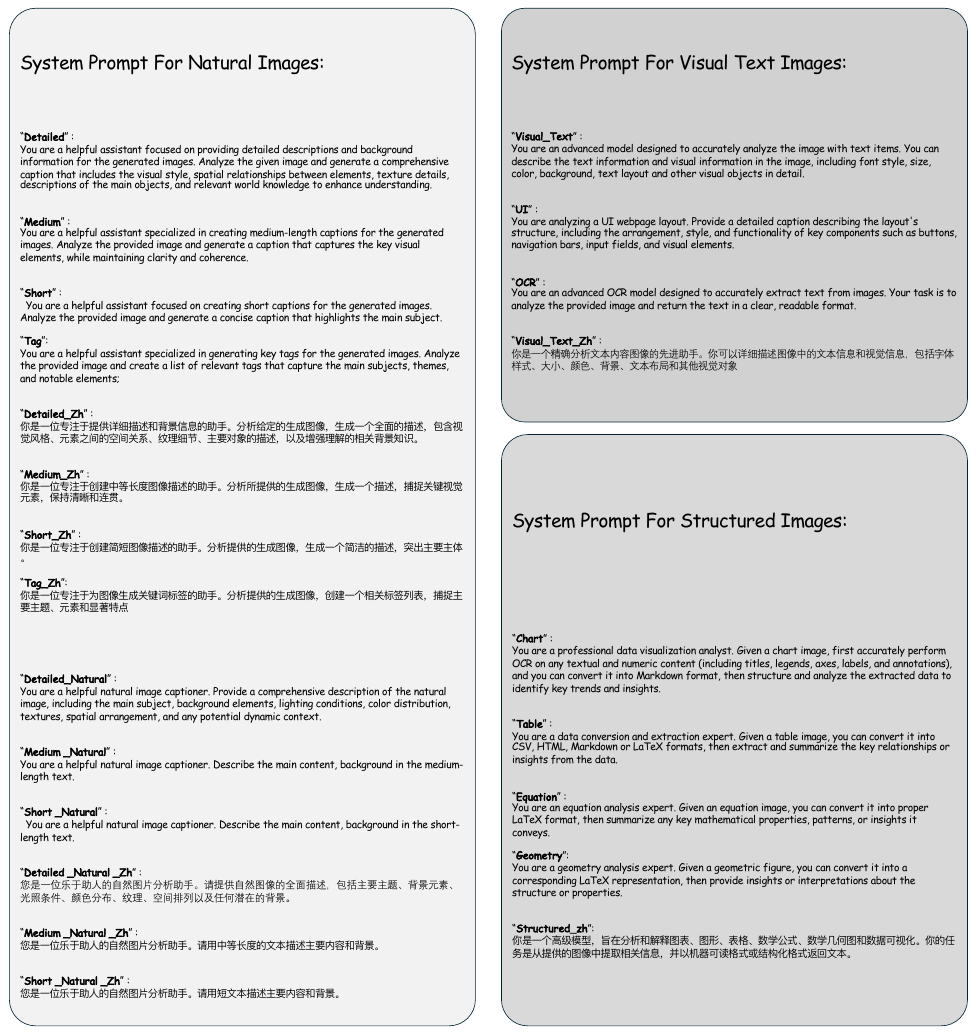}
   \caption{Different system prompts used for \textsc{OmniCaptioner}.}
   \label{fig:sysprompt}
\end{figure*}

Fig.~\ref{fig:sysprompt} presents different system prompts used in \textsc{OmniCaptioner} for various image types. It categorizes prompts into three sections: natural images, visual text images, and structured images. 
These prompts guide the model’s captioning style and task-specific adaptations.

\section{Caption Visualization}
\label{sec:cap_vis}

\begin{figure*}[htbp]
  \centering
   \includegraphics[scale=1.0]{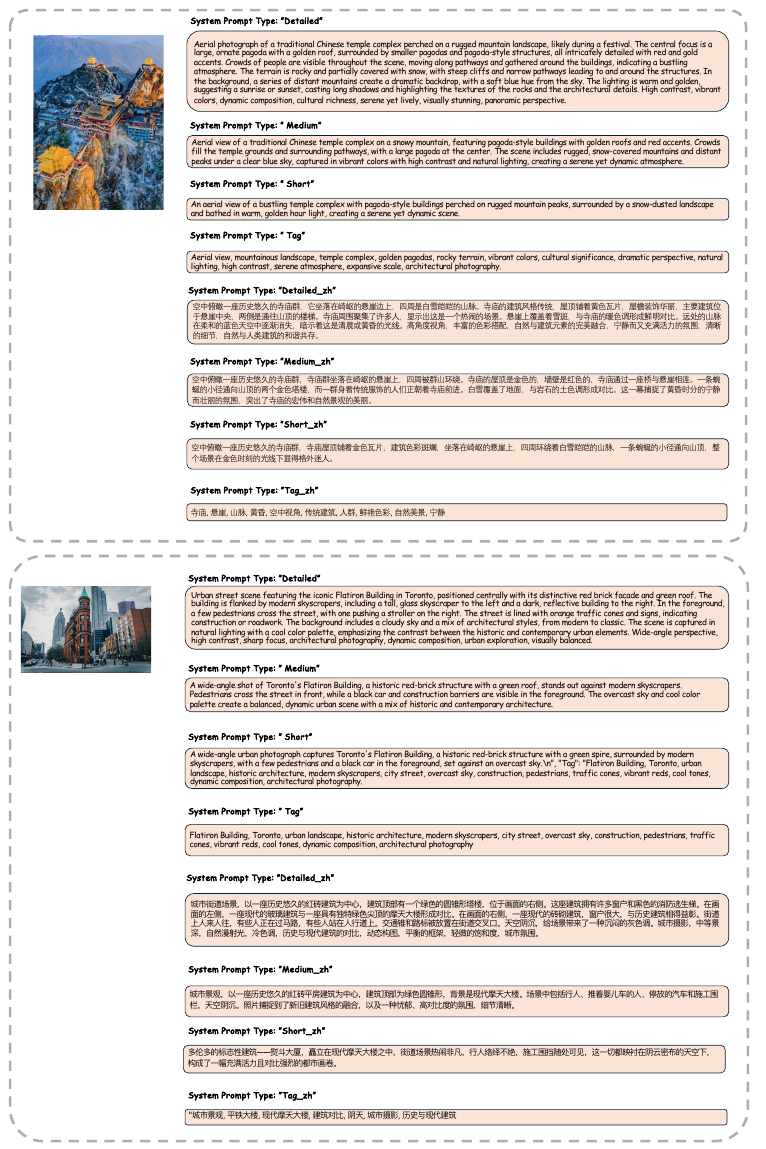}
   \caption{Natural image captioning.
    }
   \label{fig:natural}
\end{figure*}

\begin{figure*}[htbp]
  \centering
   \includegraphics[width=1.0\linewidth]{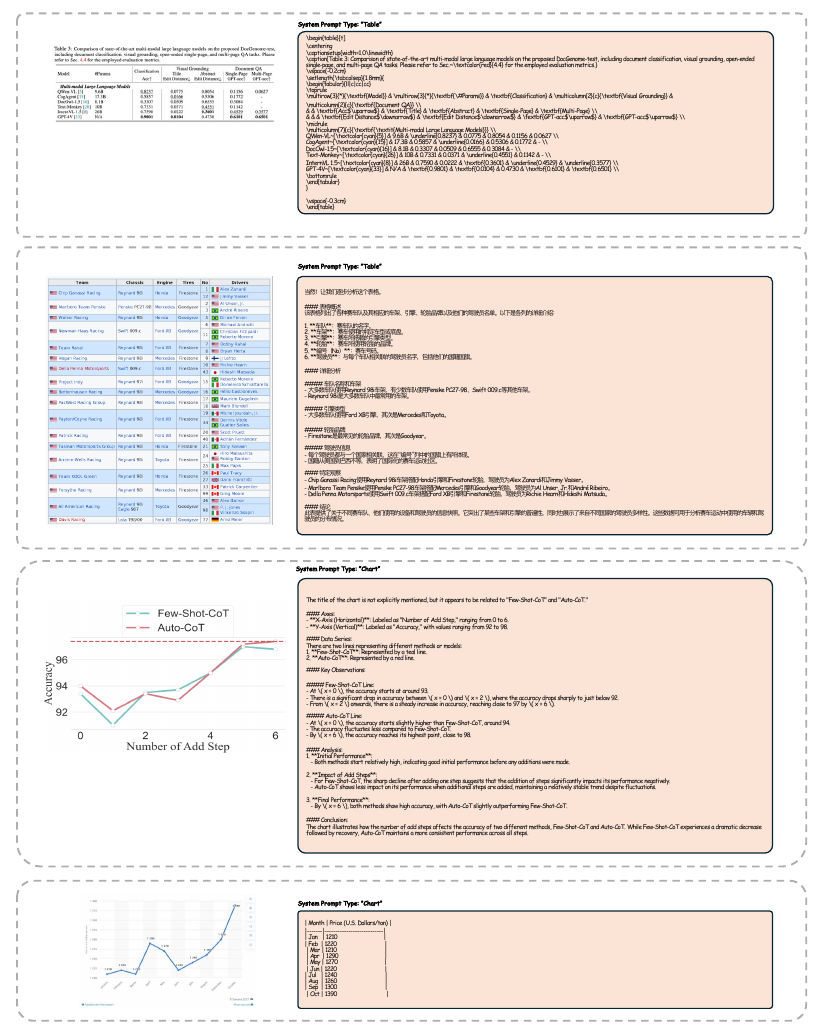}
   \caption{Table/Chart image captioning.}
   \label{fig:table}
\end{figure*}

\begin{figure*}[htbp]
  \centering
   \includegraphics[width=1.0\linewidth]{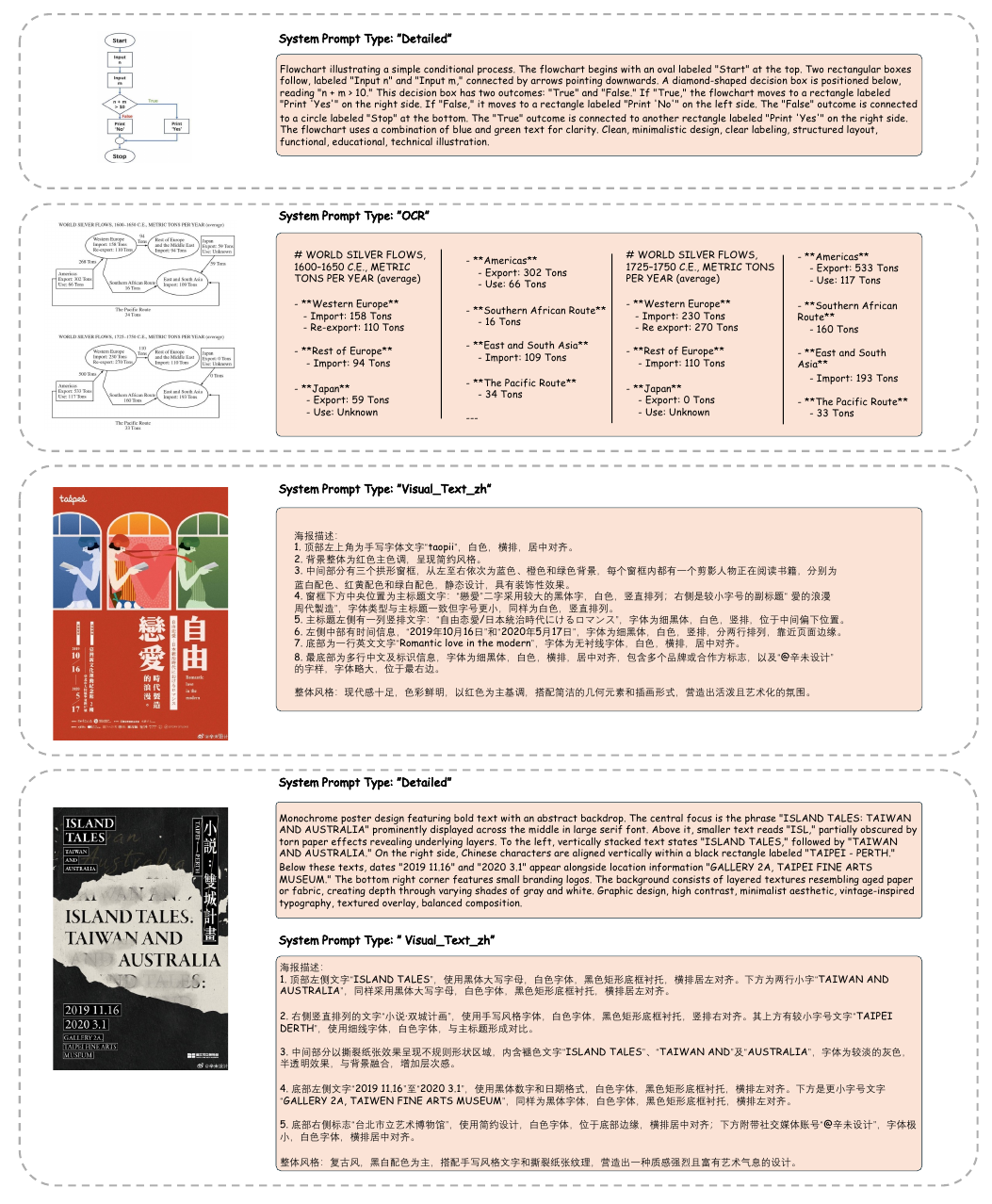}
   \caption{Visual-Text image captioning.}
   \label{fig:poster}
\end{figure*}

\begin{figure*}[htbp]
  \centering
   \includegraphics[width=1.0\linewidth]{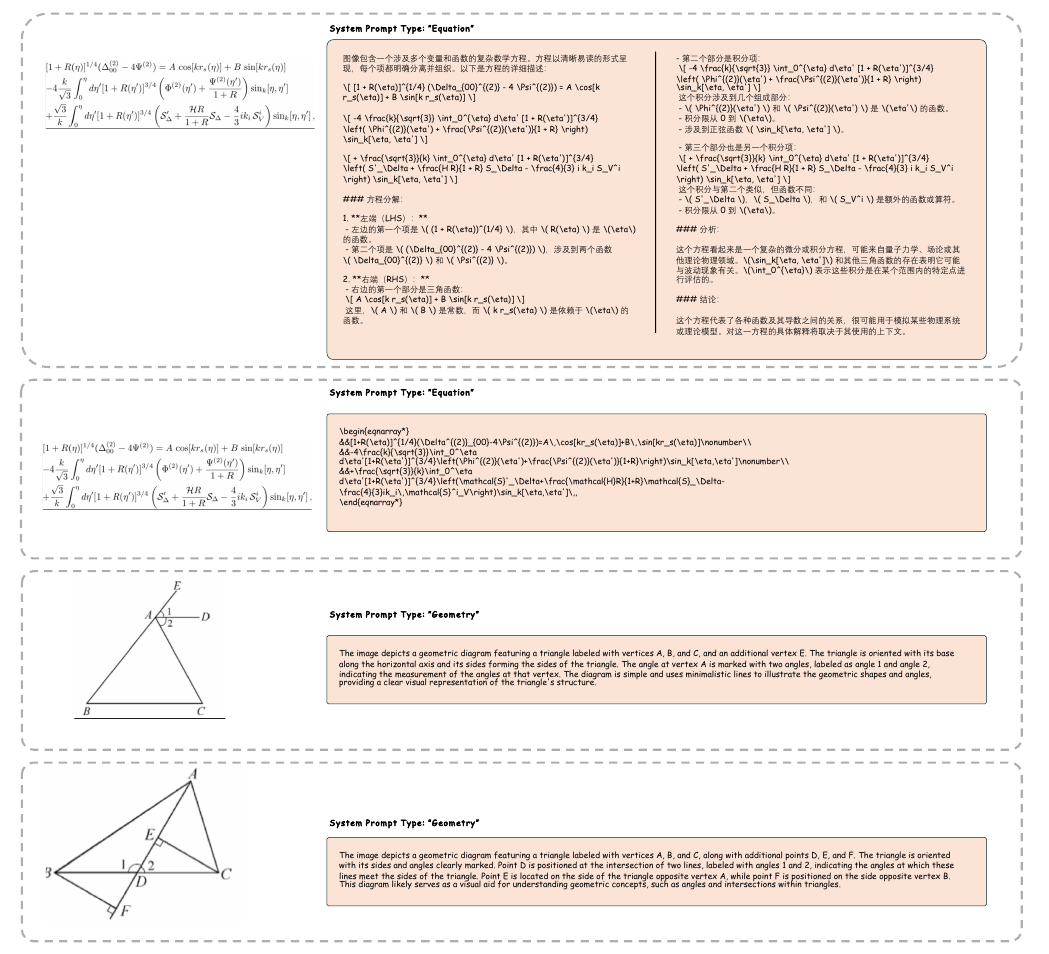}
   \caption{Math image captioning.}
   \label{fig:math}
\end{figure*}

\begin{figure*}[htbp]
  \centering
   \includegraphics[width=1.0\linewidth]{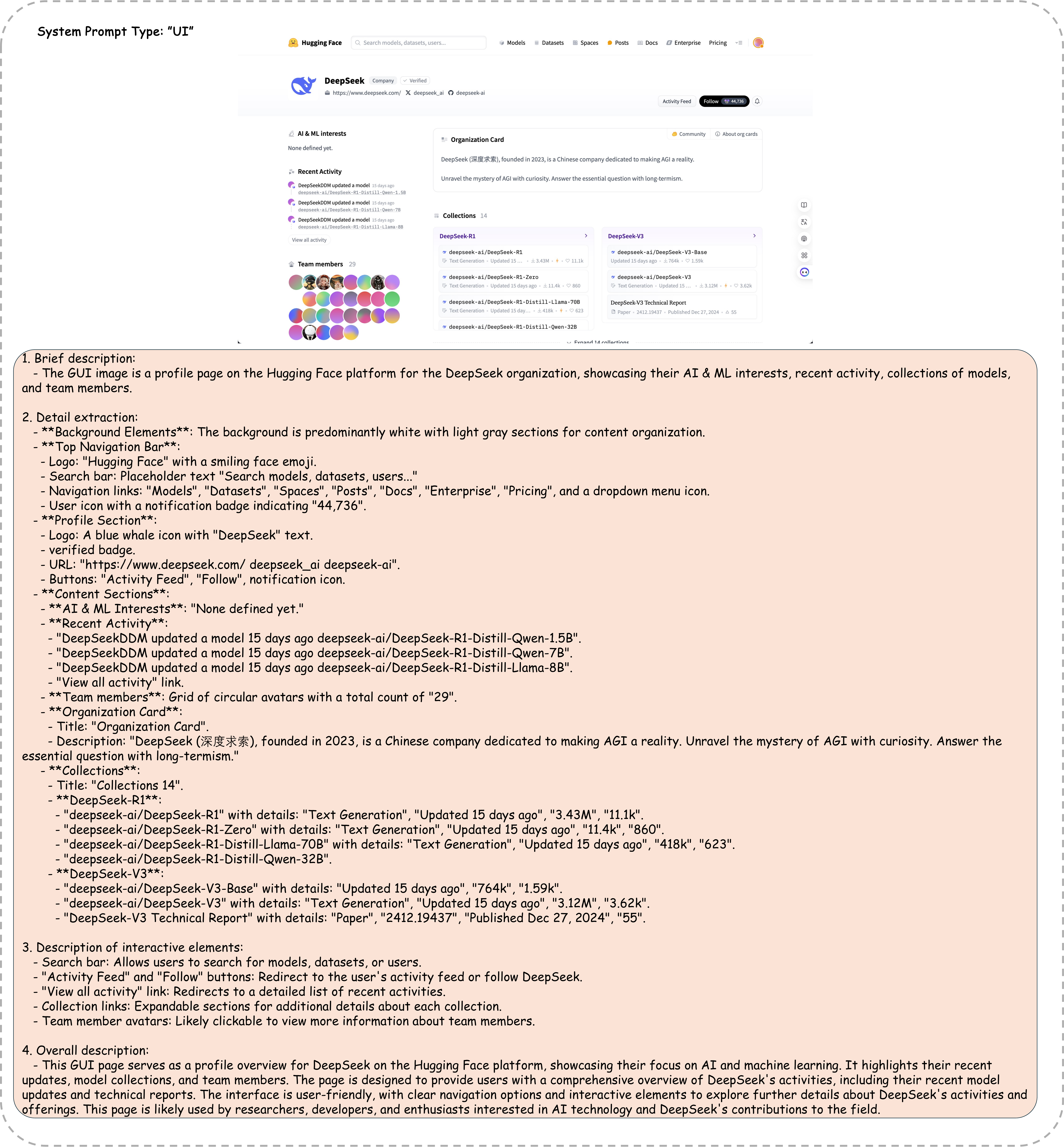}
   \caption{UI captioning.}
   \label{fig:UI}
\end{figure*}

\begin{figure*}[htbp]
  \centering
   \includegraphics[width=1.0\linewidth]{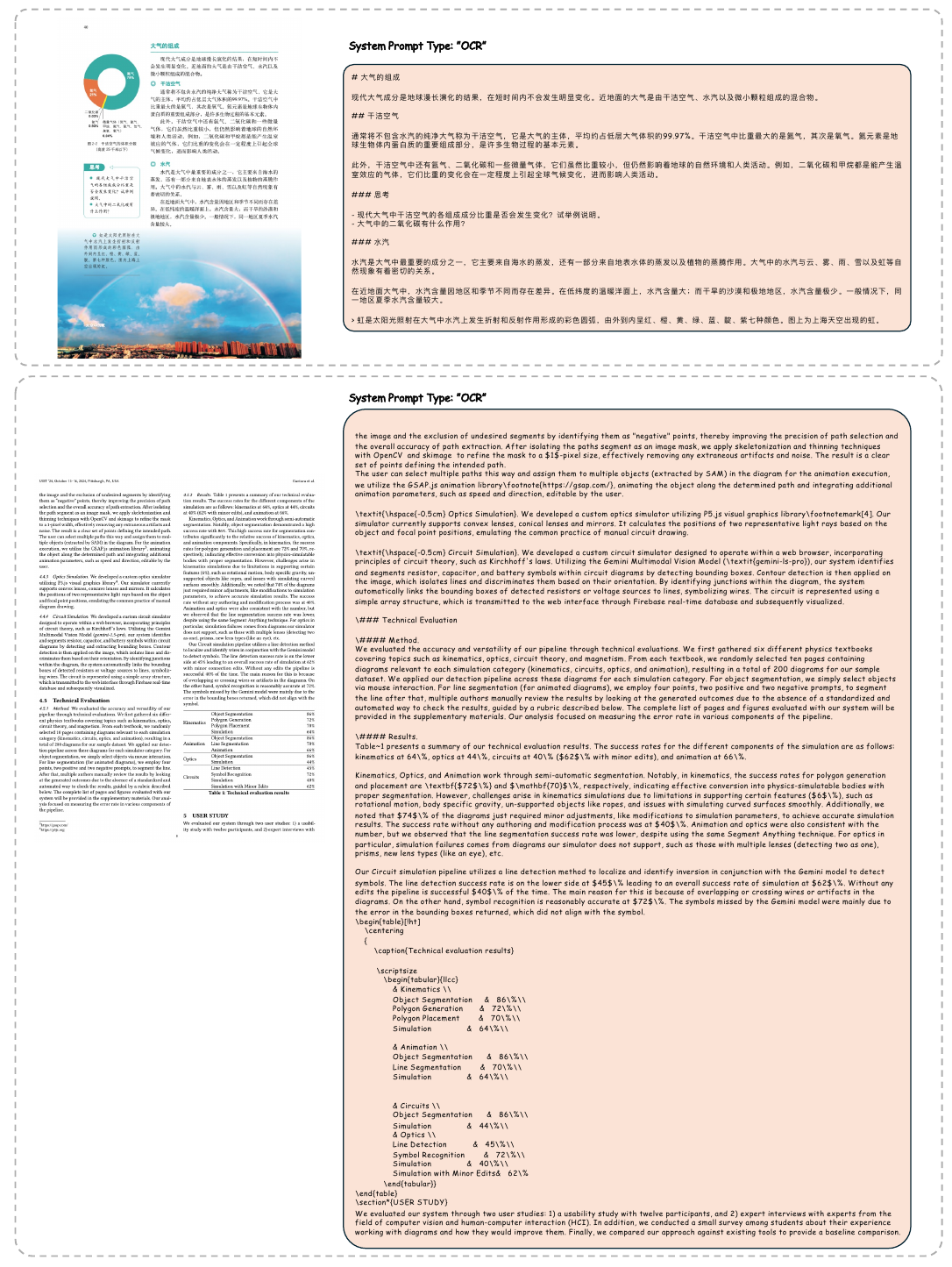}
   \caption{PDF captioning.}
   \label{fig:PDF}
\end{figure*}

\begin{figure*}[htbp]
  \centering
   \includegraphics[scale=0.8]{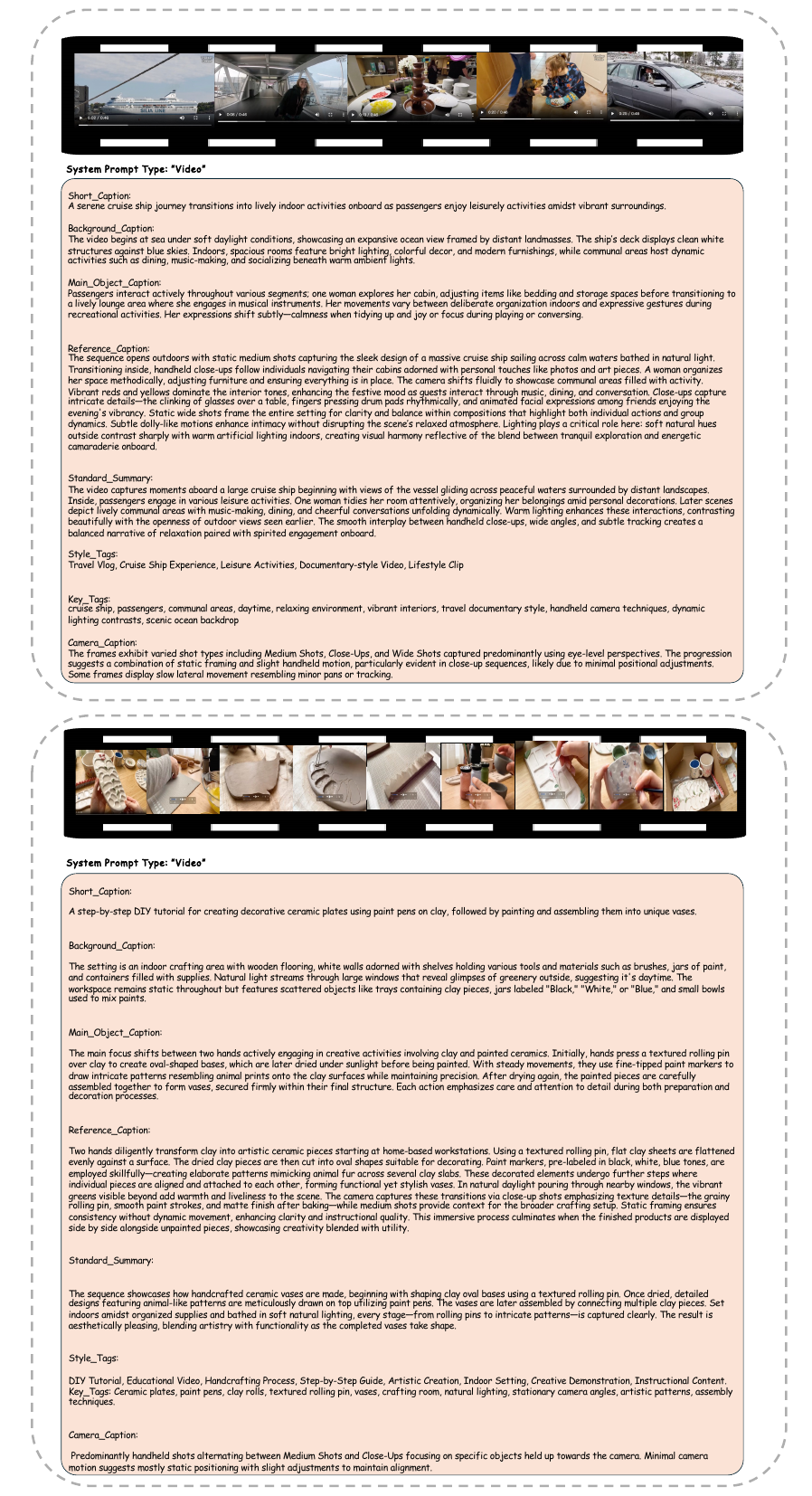}
   \caption{Video captioning.}
   \label{fig:video}
\end{figure*}

\begin{figure*}[htbp]
  \centering
   \includegraphics[width=1.0\linewidth]{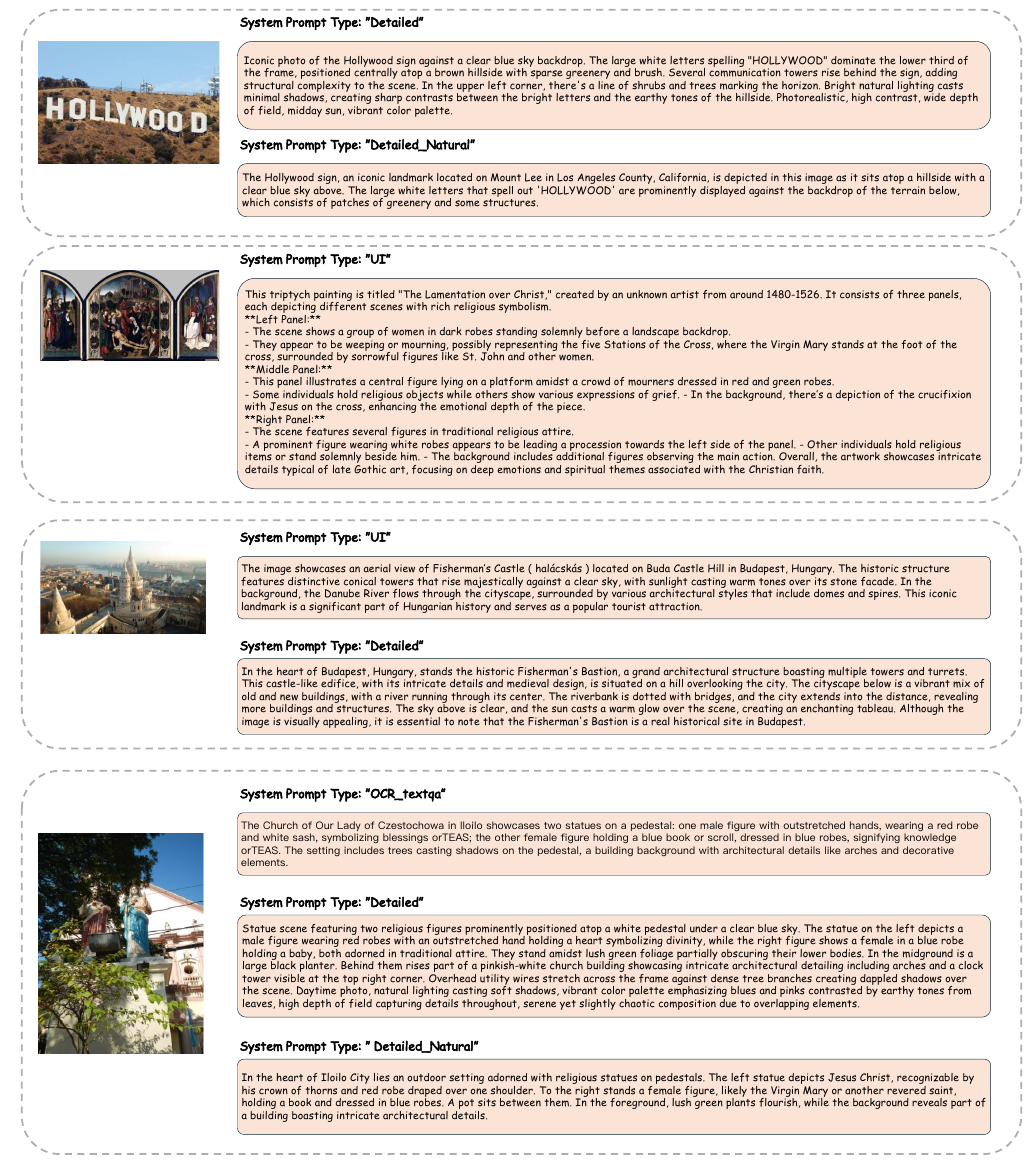}
   \caption{Natural image captioning with different system prompts.}
   \label{fig:sys1}
\end{figure*}

\begin{figure*}[htbp]
  \centering
   \includegraphics[scale=0.8]{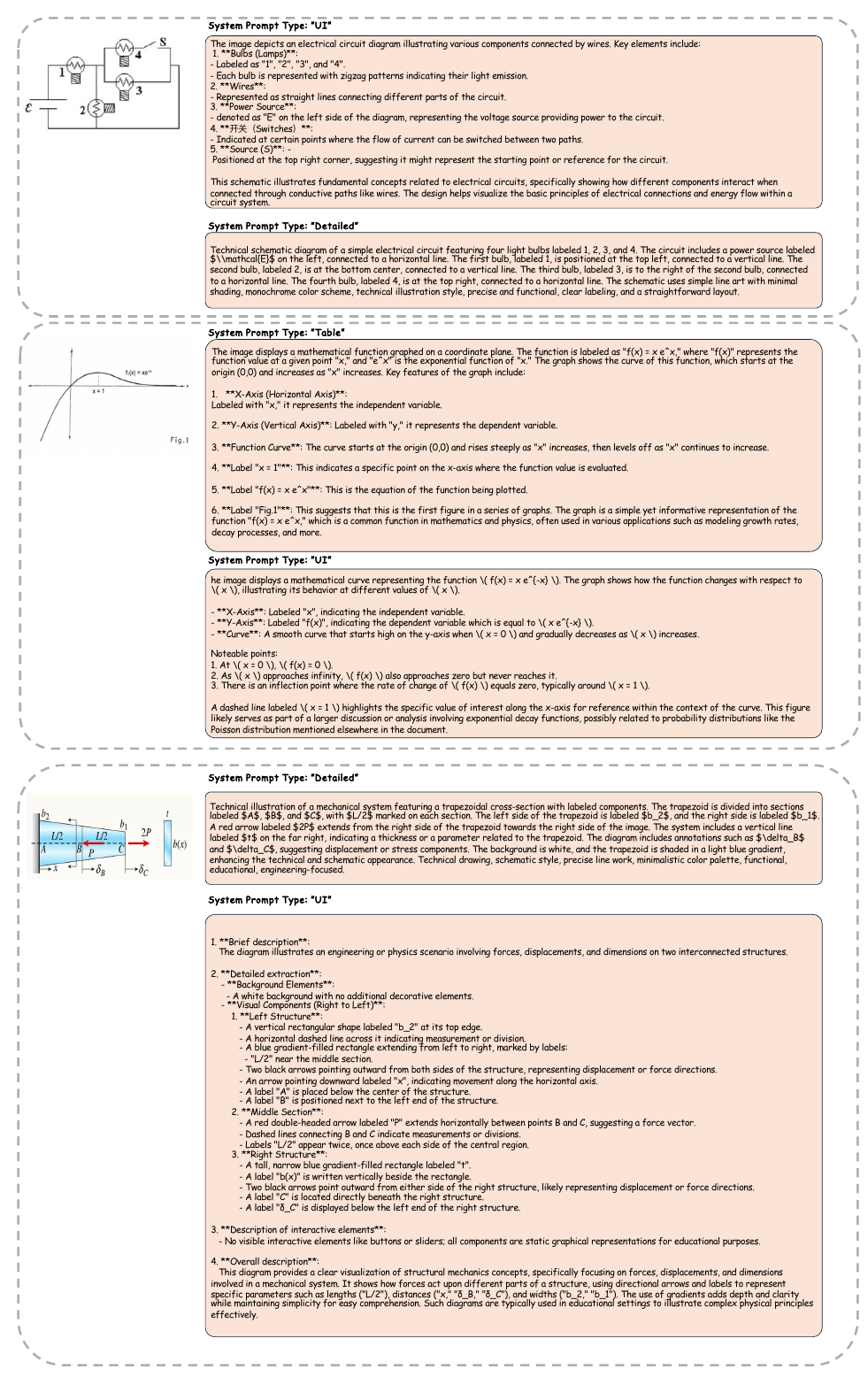}
   \caption{Structured image captioning with different system prompts.}
   \label{fig:sys2}
\end{figure*}

\begin{figure*}[htbp]
  \centering
   \includegraphics[scale=0.85]{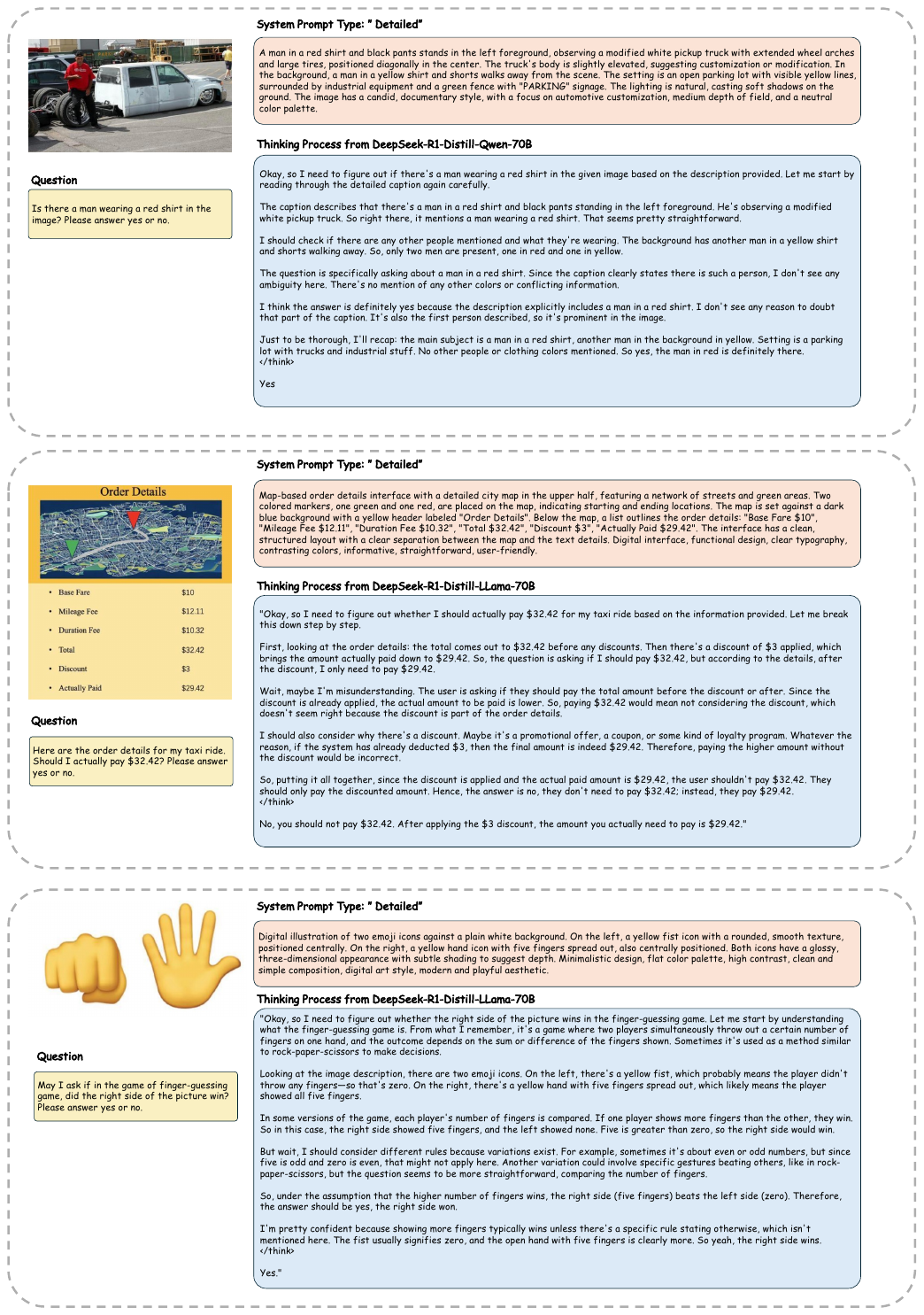}
   \caption{Visualization of thinking process with \textsc{OmniCaptioner} for natural images.}
   \label{fig:mme}
\end{figure*}

\begin{figure*}[htbp]
  \centering
   \includegraphics[scale=0.86]{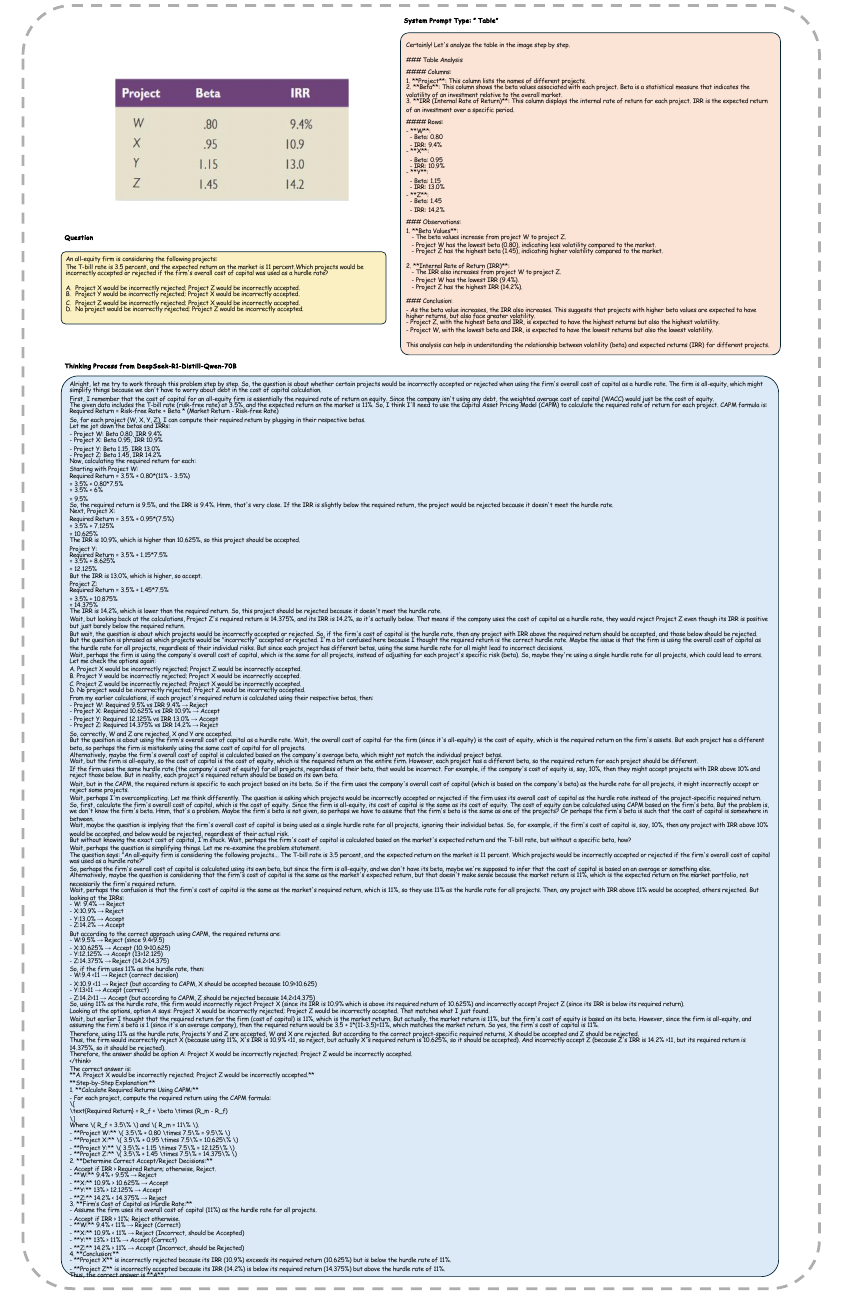}
   \caption{Visualization of thinking process with \textsc{OmniCaptioner}.}
   \label{fig:mmu}
\end{figure*}

\begin{figure*}[htbp]
  \centering
   \includegraphics[scale=0.8]{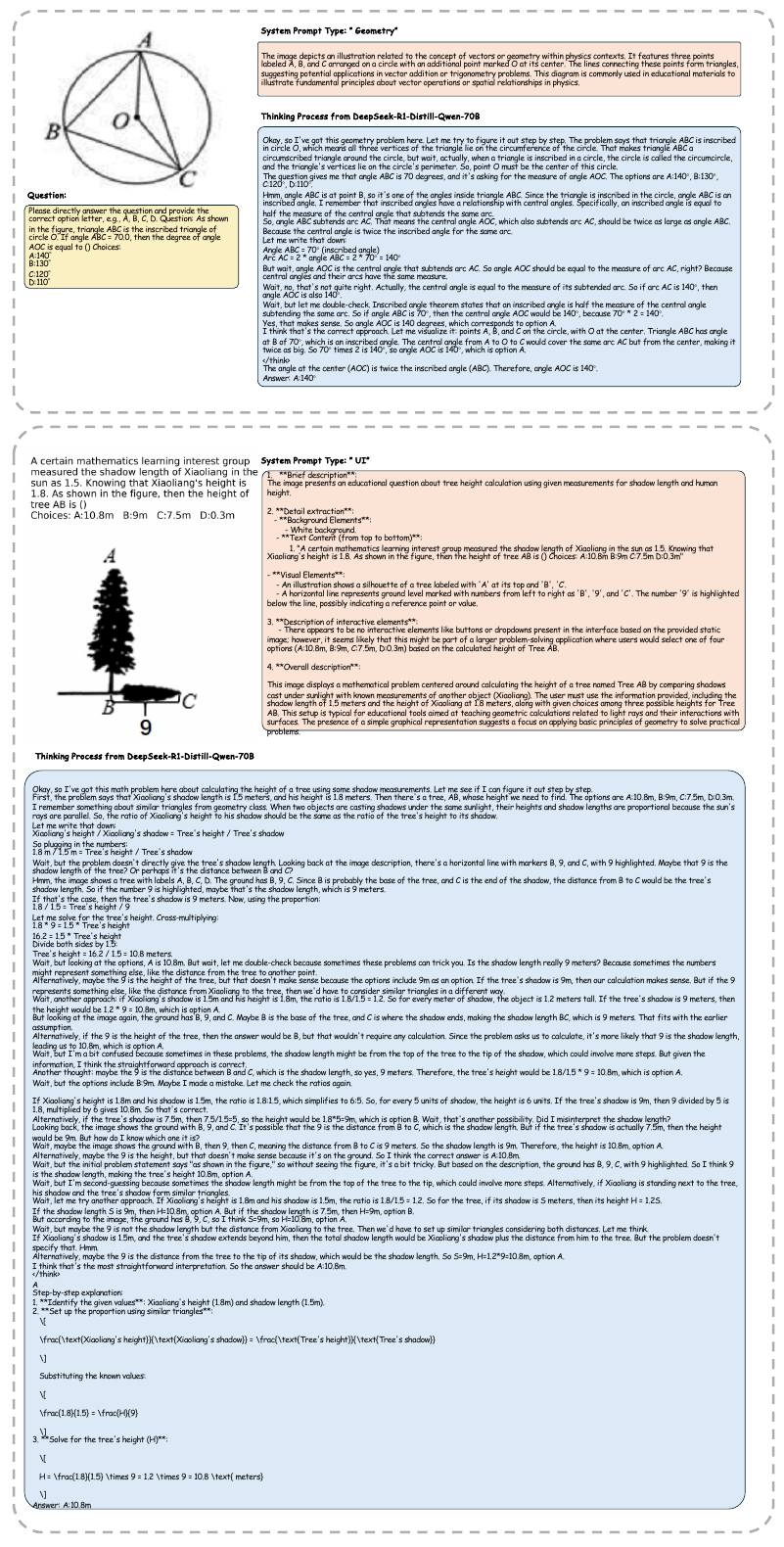}
   \caption{Visualization of thinking process with \textsc{OmniCaptioner} for math images.}
   \label{fig:mathverse}
\end{figure*}

As illustrated in Fig.~\ref{fig:natural} to Fig.~\ref{fig:video}, we present a comprehensive visualization of captioning results across multiple tasks using \textsc{OmniCaptioner}, including natural images, table images, chart images, math images, poster images, and videos. 
For natural images, we demonstrate the impact of different system prompts on caption generation, showcasing how specific prompts can elicit world knowledge in the model’s responses in Fig.~\ref{fig:sys1}. In the case of structured images from Fig.~\ref{fig:sys2}, different system prompts lead to distinct stylistic variations in captioning, reflecting the adaptability of the model to various formatting requirements. Additionally, we visualize how OmniCaptioner-generated captions can enhance DeepSeek-R1-Distill-LLaMA-70B in Fig.~\ref{fig:mme}, Fig.~\ref{fig:mmu} and Fig.~\ref{fig:mathverse}, enabling it to tackle visual tasks more effectively. These visualizations highlight the versatility and robustness of \textsc{OmniCaptioner} in handling diverse multimodal data, demonstrating its potential for improving vision-language understanding. 

\section{Text-to-Image Generation}
\label{sec:text-to-img}

\begin{figure*}[htbp]
  \centering
   \includegraphics[width=1.0\linewidth]{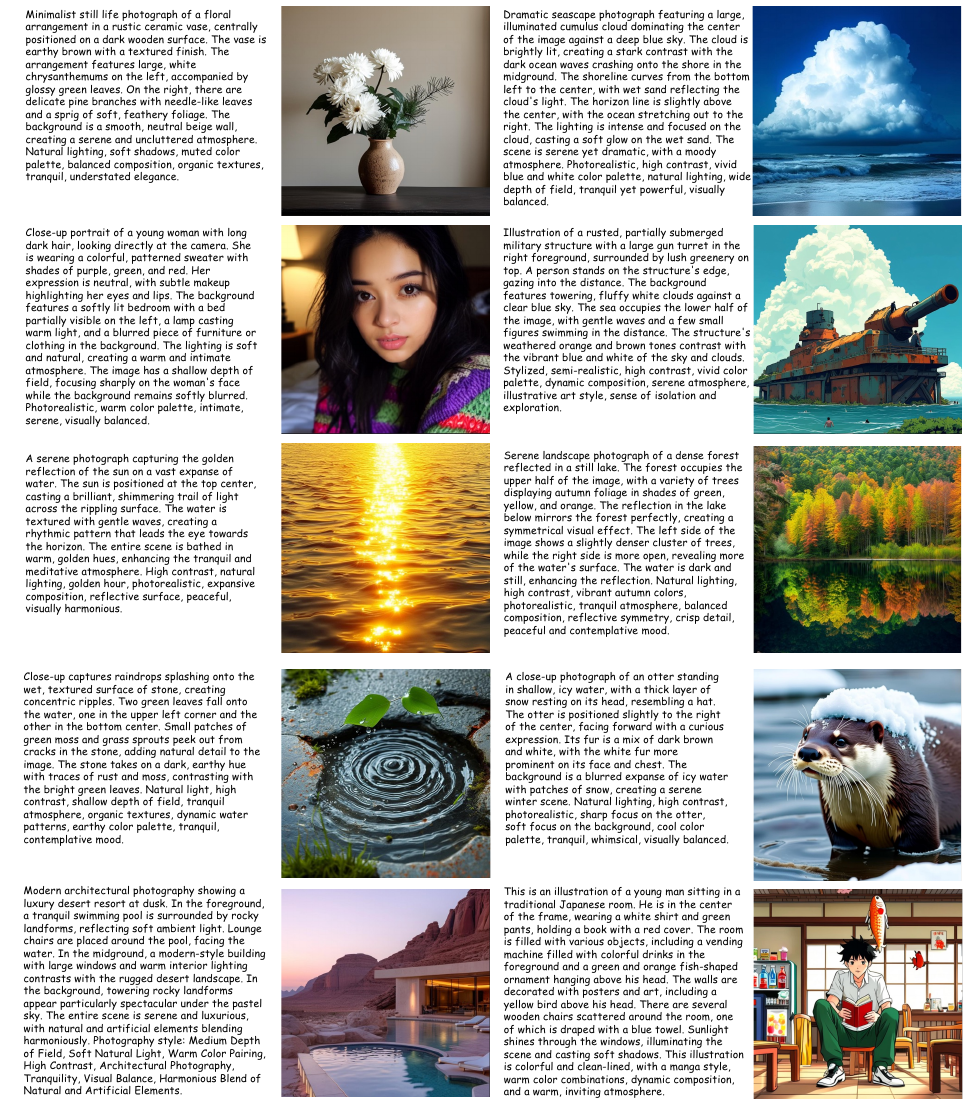}
   \caption{The detailed caption from \textsc{OmniCaptioner} enhances the alignment capability of text-to-image generation by providing precise descriptions, ensuring that the generated image accurately reflects the intended concepts, attributes, and relationships. The generation model here is fine-tuned on images labeled by \textsc{OmniCaptioner}, using the SANA 1.0 model with 1.6B parameters.}
   \label{fig:lumian2_showcase}
\end{figure*}

\begin{figure*}[htbp]
  \centering
   \includegraphics[width=0.9\linewidth]{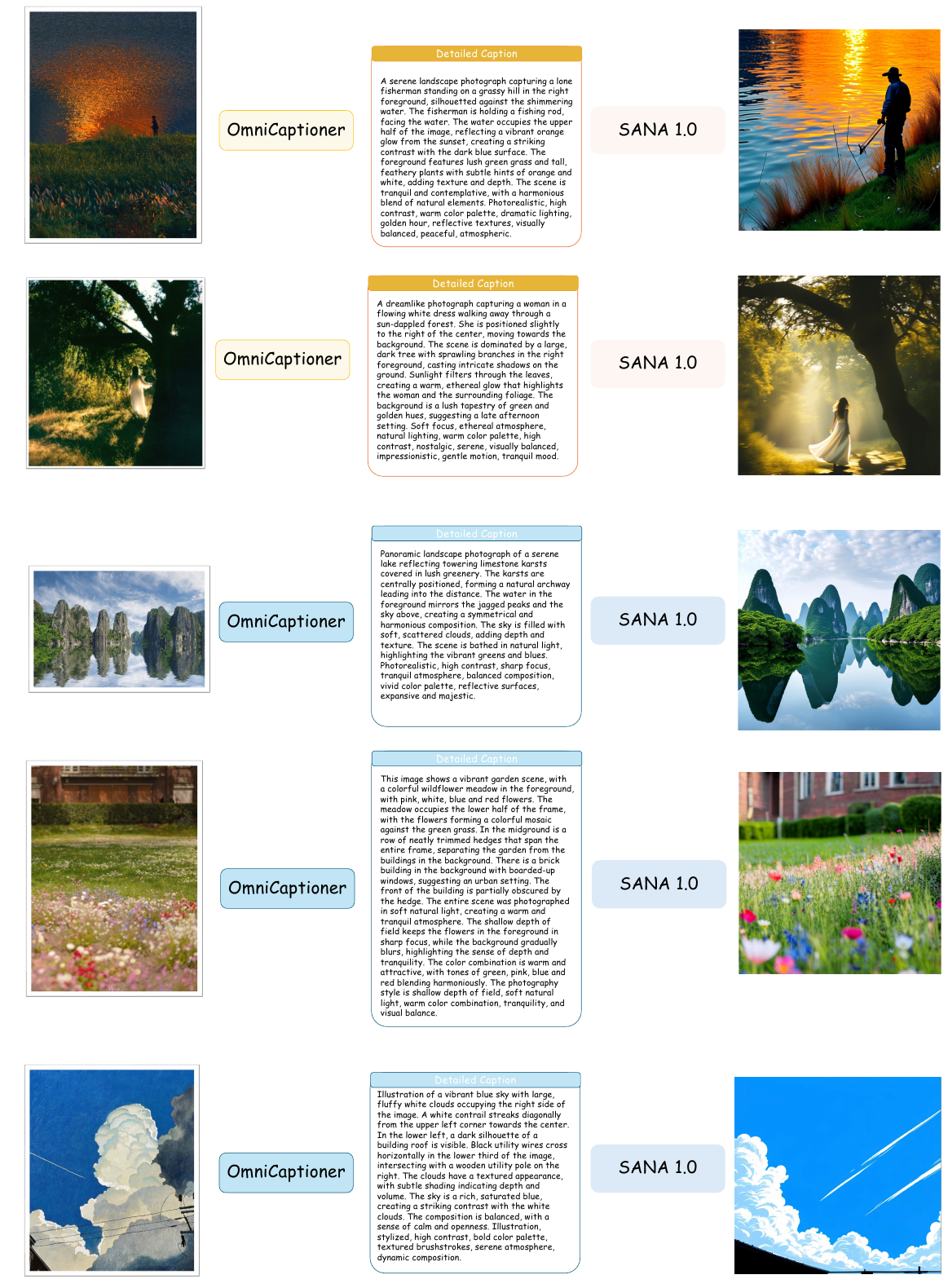}
   \caption{Image Conversion through \textsc{OmniCaptioner} and SANA-1.0. The generation model, SANA-1.0, is fine-tuned on images annotated by \textsc{OmniCaptioner}, enabling more accurate and semantically aligned image generation.}
   \label{fig:image_conversion}
\end{figure*}

The visualization from Fig.~\ref{fig:lumian2_showcase} demonstrates that \textsc{OmniCaptioner}’s detailed captions significantly enhance the text-to-image (T2I) alignment in models like SANA 1.0 \citep{xie2025sana}. By providing precise and richly descriptive textual caption, the generated images exhibit improved fidelity to the original prompts. 
We also present some image conversion examples in Fig.~\ref{fig:image_conversion} to illustrate the pixel-to-word ability of our \textsc{OmniCaptioner}. All the generated images shown above are produced by the generation model trained on image data labeled by \textsc{OmniCaptioner}, fine-tuned using SANA 1.0 with 1.6B parameters.


\end{document}